\documentclass{article} %
\usepackage{times}
\usepackage{graphicx}
\usepackage{subfigure}
\usepackage{natbib}
\usepackage{algorithm}
\usepackage{algpseudocode}
\usepackage{hyperref}

\usepackage[accepted]{icml2016}
\usepackage{url}
\usepackage{amsfonts}
\usepackage{amsmath}
\usepackage{color}
\usepackage{afterpage}
\usepackage{xr}
\usepackage[T1]{fontenc}

\newcommand{\vlad}{Volodymyr Mnih}
\newcommand{\adria}{Adri\`a Puigdom\`enech Badia}
\newcommand{\mehdi}{Mehdi Mirza}
\newcommand{\alex}{Alex Graves}
\newcommand{\timl}{Timothy P. Lillicrap}
\newcommand{\timh}{Tim Harley}
\newcommand{\dave}{David Silver}
\newcommand{\koray}{Koray Kavukcuoglu }

\icmltitlerunning{Asynchronous Methods for Deep Reinforcement Learning}

\begin{document}
\twocolumn[
\icmltitle{Asynchronous Methods for Deep Reinforcement Learning}

\icmlauthor{\vlad$^1$}{vmnih@google.com}
\icmlauthor{\adria$^1$}{adriap@google.com}
\icmlauthor{\mehdi$^{1,2}$}{mirzamom@iro.umontreal.ca}
\icmlauthor{\alex$^1$}{gravesa@google.com}
\icmlauthor{\timh$^1$}{tharley@google.com}
\icmlauthor{\timl$^1$}{countzero@google.com}
\icmlauthor{\dave$^1$}{davidsilver@google.com}
\icmlauthor{\koray$^1$}{korayk@google.com}
\icmladdress{$^1$ Google DeepMind \\ $^2$ Montreal Institute for Learning Algorithms (MILA), University of Montreal \\}

]

\begin{abstract}

We propose a conceptually simple and lightweight framework for deep reinforcement learning that uses asynchronous gradient descent for optimization of deep neural network controllers.
We present asynchronous variants of four standard reinforcement learning algorithms and show that parallel actor-learners have a stabilizing effect on training allowing all four methods to successfully train neural network controllers.
The best performing method, an asynchronous variant of actor-critic, surpasses the current state-of-the-art on the Atari domain while training for half the time on a single multi-core CPU instead of a GPU.
Furthermore, we show that asynchronous actor-critic succeeds on a wide variety of continuous motor control problems as well as on a new task of navigating random 3D mazes using a visual input.

\end{abstract}

\vspace{-0.2cm}
\vspace{-0.1cm}
\section{Introduction}
\label{sec:intro}

Deep neural networks provide rich representations that can enable reinforcement learning (RL) algorithms to perform effectively.
However, it was previously thought that the combination of simple online RL algorithms with deep neural networks was fundamentally unstable.
Instead, a variety of solutions have been proposed to stabilize the algorithm \citep{riedmiller2005nfq,mnih-atari-2013,mnih-dqn-2015,hado2015doubledqn,schulman2015trust}.
These approaches share a common idea: the sequence of observed data encountered by an online RL agent is non-stationary, and online RL updates are strongly correlated.
By storing the agent's data in an experience replay memory, the data can be batched \citep{riedmiller2005nfq,schulman2015trust} or randomly sampled~\citep{mnih-atari-2013,mnih-dqn-2015,hado2015doubledqn} from different time-steps.
Aggregating over memory in this way reduces non-stationarity and decorrelates updates, but at the same time limits the methods to off-policy reinforcement learning algorithms.

Deep RL algorithms based on experience replay have achieved unprecedented success in challenging domains such as Atari 2600.
However, experience replay has several drawbacks: it uses more memory and computation per real interaction; and it requires off-policy learning algorithms that can update from data generated by an older policy.

In this paper we provide a very different paradigm for deep reinforcement learning.
Instead of experience replay, we asynchronously execute multiple agents in parallel, on multiple instances of the environment.
This parallelism also decorrelates the agents' data into a more stationary process, since at any given time-step the parallel agents will be experiencing a variety of different states.
This simple idea enables a much larger spectrum of fundamental on-policy RL algorithms, such as Sarsa, n-step methods, and actor-critic methods, as well as off-policy RL algorithms such as Q-learning, to be applied robustly and effectively using deep neural networks.

Our parallel reinforcement learning paradigm also offers practical benefits.
Whereas previous approaches to deep reinforcement learning rely heavily on specialized hardware such as GPUs \citep{mnih-dqn-2015,hado2015doubledqn,schaul2015prioritized} or massively distributed architectures \citep{nair2015gorila}, our experiments run on a single machine with a standard multi-core CPU.
When applied to a variety of Atari 2600 domains, on many games asynchronous reinforcement learning achieves better results, in far less time than previous GPU-based algorithms, using far less resource than massively distributed approaches.
The best of the proposed methods, asynchronous advantage actor-critic (A3C), also mastered a variety of continuous motor control tasks as well as learned general strategies for exploring 3D mazes purely from visual inputs.
We believe that the success of A3C on both 2D and 3D games, discrete and continuous action spaces, as well as its ability to train feedforward and recurrent agents makes it the most general and successful reinforcement learning agent to date.

\vspace{-0.1cm}
\section{Related Work}
\label{sec:related}

The General Reinforcement Learning Architecture (Gorila) of~\cite{nair2015gorila} performs asynchronous training of reinforcement learning agents in a distributed setting.
In Gorila, each process contains an actor that acts in its own copy of the environment, a separate replay memory, and a learner that samples data from the replay memory and computes gradients of the DQN loss~\citep{mnih-dqn-2015} with respect to the policy parameters.
The gradients are asynchronously sent to a central parameter server which updates a central copy of the model.
The updated policy parameters are sent to the actor-learners at fixed intervals.
By using 100 separate actor-learner processes and 30 parameter server instances, a total of 130 machines, Gorila was able to significantly outperform DQN over 49 Atari games.
On many games Gorila reached the score achieved by DQN over 20 times faster than DQN.
We also note that a similar way of parallelizing DQN was proposed by~\cite{chavez2015deepq}.

In earlier work, \cite{li:mapreduce} applied the Map Reduce framework to parallelizing batch reinforcement learning methods with linear function approximation.
Parallelism was used to speed up large matrix operations but not to parallelize the collection of experience or stabilize learning.
\cite{grounds:parallelRL} proposed a parallel version of the Sarsa algorithm that uses multiple separate actor-learners to accelerate training.
Each actor-learner learns separately and periodically sends updates to weights that have changed significantly to the other learners using peer-to-peer communication.

\cite{tsitsiklis1994asynchronous} studied convergence properties of Q-learning in the asynchronous optimization setting.
These results show that Q-learning is still guaranteed to converge when some of the information is outdated as long as outdated information is always eventually discarded and several other technical assumptions are satisfied.
Even earlier, \cite{bertsekas1982distributed} studied the related problem of distributed dynamic programming.

Another related area of work is in evolutionary methods, which are often straightforward to parallelize by distributing fitness evaluations over multiple machines or threads~\citep{tomassini1999parallel}.
Such parallel evolutionary approaches have recently been applied to some visual reinforcement learning tasks.
In one example, \cite{koutnik2014} evolved convolutional neural network controllers for the TORCS driving simulator by performing fitness evaluations on 8 CPU cores in parallel.

\vspace{-0.1cm}
\section{Reinforcement Learning Background}
\label{sec:background}

We consider the standard reinforcement learning setting where an agent interacts with an environment $\mathcal{E}$ over a number of discrete time steps.
At each time step $t$, the agent receives a state $s_t$ and selects an action $a_t$ from some set of possible actions $\mathcal{A}$ according to its policy $\pi$, where $\pi$ is a mapping from states $s_t$ to actions $a_t$.
In return, the agent receives the next state $s_{t+1}$ and receives a scalar reward $r_t$.
The process continues until the agent reaches a terminal state after which the process restarts.
The return $R_t = \sum_{k=0}^{\infty} \gamma^k r_{t+k}$ is the total accumulated return from time step $t$ with discount factor $\gamma \in (0,1]$.
The goal of the agent is to maximize the expected return from each state $s_t$.

The action value $Q^{\pi}(s,a) = \mathbb{E}\left[R_t|s_t=s, a\right]$ is the expected return for selecting action $a$ in state $s$ and following policy $\pi$.
The optimal value function $Q^*(s,a) = \max_{\pi} Q^{\pi}(s,a)$ gives the maximum action value for state $s$ and action $a$ achievable by any policy.
Similarly, the value of state $s$ under policy $\pi$ is defined as $V^{\pi}(s) = \mathbb{E}\left[R_t|s_t=s\right]$ and is simply the expected return for following policy $\pi$ from state $s$.

In value-based model-free reinforcement learning methods, the action value function is represented using a function approximator, such as a neural network.
Let $Q(s,a;\theta)$ be an approximate action-value function with parameters $\theta$.
The updates to $\theta$ can be derived from a variety of reinforcement learning algorithms.
One example of such an algorithm is Q-learning, which aims to directly approximate the optimal action value function: $Q^*(s,a)\approx Q(s,a;\theta)$.
In one-step Q-learning, the parameters $\theta$ of the action value function $Q(s,a;\theta)$ are learned by iteratively minimizing a sequence of loss functions, where the $i$th loss function defined as
\begin{equation*}
    L_i(\theta_i) = \mathbb{E}\left(r + \gamma \max_{a'}Q(s', a';\theta_{i-1}) - Q(s,a;\theta_i) \right)^2
\end{equation*}
where $s'$ is the state encountered after state $s$.

We refer to the above method as one-step Q-learning because it updates the action value $Q(s,a)$ toward the one-step return $r+\gamma \max_{a'}Q(s',a';\theta)$.
One drawback of using one-step methods is that obtaining a reward $r$ only directly affects the value of the state action pair $s,a$ that led to the reward.
The values of other state action pairs are affected only indirectly through the updated value $Q(s,a)$.
This can make the learning process slow since many updates are required the propagate a reward to the relevant preceding states and actions.

One way of propagating rewards faster is by using $n$-step returns~\citep{watkins1989learning,peng1996msq}. In $n$-step Q-learning, $Q(s,a)$ is updated toward the $n$-step return defined as
    $r_t + \gamma r_{t+1} + \cdots + \gamma^{n-1} r_{t+n-1} + \max_a \gamma^n Q(s_{t+n}, a)$.
This results in a single reward $r$ directly affecting the values of $n$ preceding state action pairs.
This makes the process of propagating rewards to relevant state-action pairs potentially much more efficient.

In contrast to value-based methods, policy-based model-free methods directly parameterize the policy $\pi(a|s;\theta)$ and update the parameters $\theta$ by performing, typically approximate, gradient ascent on $\mathbb{E}[R_t]$.
One example of such a method is the REINFORCE family of algorithms due to Williams~\yrcite{Williams1992}.
Standard REINFORCE updates the policy parameters $\theta$ in the direction $\nabla_{\theta}\log\pi(a_t|s_t;\theta) R_t$, which is an unbiased estimate of $\nabla_{\theta} \mathbb{E}[R_t]$.
It is possible to reduce the variance of this estimate while keeping it unbiased by subtracting a learned function of the state $b_t(s_t)$, known as a baseline~\citep{Williams1992}, from the return.
The resulting gradient is
$\nabla_{\theta}\log\pi(a_t|s_t;\theta) \left(R_t-b_t(s_t)\right)$.

A learned estimate of the value function is commonly used as the baseline $b_t(s_t)\approx V^{\pi}(s_t)$ leading to a much lower variance estimate of the policy gradient.
When an approximate value function is used as the baseline, the quantity $R_t-b_t$ used to scale the policy gradient can be seen as an estimate of the \emph{advantage} of action $a_t$ in state $s_t$, or $A(a_t,s_t)=Q(a_t,s_t)-V(s_t)$, because $R_t$ is an estimate of $Q^{\pi}(a_t,s_t)$ and $b_t$ is an estimate of $V^{\pi}(s_t)$.
This approach can be viewed as an actor-critic architecture where the policy $\pi$ is the actor and the baseline $b_t$ is the critic~\citep{sutton:book,degris2012model}.

\vspace{-0.1cm}
\section{Asynchronous RL Framework}
\label{sec:method}

We now present multi-threaded asynchronous variants of one-step Sarsa, one-step Q-learning, n-step Q-learning, and advantage actor-critic.
The aim in designing these methods was to find RL algorithms that can train deep neural network policies reliably and without large resource requirements.
While the underlying RL methods are quite different, with actor-critic being an on-policy policy search method and Q-learning being an off-policy value-based method, we use two main ideas to make all four algorithms practical given our design goal.

First, we use asynchronous actor-learners, similarly to the Gorila framework~\citep{nair2015gorila}, but instead of using separate machines and a parameter server, we use multiple CPU threads on a single machine.
Keeping the learners on a single machine removes the communication costs of sending gradients and parameters and enables us to use Hogwild!~\citep{recht2011hogwild} style updates for training.

Second, we make the observation that multiple actors-learners running in parallel are likely to be exploring different parts of the environment.
Moreover, one can explicitly use different exploration policies in each actor-learner to maximize this diversity.
By running different exploration policies in different threads, the overall changes being made to the parameters by multiple actor-learners applying online updates in parallel are likely to be less correlated in time than a single agent applying online updates.
Hence, we do not use a replay memory and rely on parallel actors employing different exploration policies to perform the stabilizing role undertaken by experience replay in the DQN training algorithm.

In addition to stabilizing learning, using multiple parallel actor-learners has multiple practical benefits.
First, we obtain a reduction in training time that is roughly linear in the number of parallel actor-learners.
Second, since we no longer rely on experience replay for stabilizing learning we are able to use on-policy reinforcement learning methods such as Sarsa and actor-critic to train neural networks in a stable way.
We now describe our variants of one-step Q-learning, one-step Sarsa, n-step Q-learning and advantage actor-critic.

\begin{algorithm}[t]
\caption{Asynchronous one-step Q-learning - pseudocode for each actor-learner thread.}
\begin{algorithmic}
\small
\State \emph{// Assume global shared $\theta$, $\theta^-$, and counter $T=0$.}
\State Initialize thread step counter $t\gets 0$
\State Initialize target network weights $\theta^- \gets \theta$
\State Initialize network gradients $d\theta \gets 0$
\State Get initial state $s$
\Repeat
\State Take action $a$ with $\epsilon$-greedy policy based on $Q(s,a;\theta)$
\State Receive new state $s'$ and reward $r$
\State $y =
    \left\{
    \begin{array}{l l}
      r  \quad & \text{for terminal } s'\\
      r + \gamma \max_{a'} Q(s', a'; \theta^-) \quad & \text{for non-terminal } s'
    \end{array} \right.$
\State Accumulate gradients wrt $\theta$: $d\theta \gets d\theta + \frac{\partial\left(y - Q(s,a;\theta)\right)^2}{\partial \theta}$
\State $s=s'$
\State $T \gets T + 1$ and $t \gets t + 1$
\If {$T\mod I_{target} == 0$}
\State Update the target network $\theta^- \gets \theta$ 
\EndIf
\If {$t\mod I_{AsyncUpdate} == 0$ or $s$ is terminal}
\State Perform asynchronous update of $\theta$ using $d\theta$.
\State Clear gradients $d\theta \gets 0$.
\EndIf
\Until $T > T_{max}$
\end{algorithmic}
\label{alg:targetq}
\end{algorithm}

\textbf{Asynchronous one-step Q-learning:}
Pseudocode for our variant of Q-learning, which we call Asynchronous one-step Q-learning, is shown in Algorithm~\ref{alg:targetq}.
Each thread interacts with its own copy of the environment and at each step computes a gradient of the Q-learning loss.
We use a shared and slowly changing target network in computing the Q-learning loss, as was proposed in the DQN training method.
We also accumulate gradients over multiple timesteps before they are applied, which is similar to using minibatches.
This reduces the chances of multiple actor learners overwriting each other's updates.
Accumulating updates over several steps also provides some ability to trade off computational efficiency for data efficiency.

Finally, we found that giving each thread a different exploration policy helps improve robustness.
Adding diversity to exploration in this manner also generally improves performance through better exploration.
While there are many possible ways of making the exploration policies differ we experiment with using $\epsilon$-greedy exploration with $\epsilon$ periodically sampled from some distribution by each thread.

\textbf{Asynchronous one-step Sarsa:}
The asynchronous one-step Sarsa algorithm is the same as asynchronous one-step Q-learning as given in Algorithm~\ref{alg:targetq} except that it uses a different target value for $Q(s,a)$.
The target value used by one-step Sarsa is $r + \gamma Q(s', a'; \theta^-)$ where $a'$ is the action taken in state $s'$~\citep{rummery1994sarsa,sutton:book}.
We again use a target network and updates accumulated over multiple timesteps to stabilize learning.

\textbf{Asynchronous n-step Q-learning:}
Pseudocode for our variant of multi-step Q-learning is shown in Supplementary Algorithm~\ref{alg:msq}.
The algorithm is somewhat unusual because it operates in the forward view by explicitly computing n-step returns, as opposed to the more common backward view used by techniques like eligibility traces~\citep{sutton:book}.
We found that using the forward view is easier when training neural networks with momentum-based methods and backpropagation through time.
In order to compute a single update, the algorithm first selects actions using its exploration policy for up to $t_{max}$ steps or until a terminal state is reached.
This process results in the agent receiving up to $t_{max}$ rewards from the environment since its last update.
The algorithm then computes gradients for n-step Q-learning updates for each of the state-action pairs encountered since the last update.
Each n-step update uses the longest possible n-step return resulting in a one-step update for the last state, a two-step update for the second last state, and so on for a total of up to $t_{max}$ updates.
The accumulated updates are applied in a single gradient step.

\textbf{Asynchronous advantage actor-critic:}
The algorithm, which we call asynchronous advantage actor-critic (A3C), maintains a policy $\pi(a_t|s_t;\theta)$ and an estimate of the value function $V(s_t;\theta_v)$.
Like our variant of n-step Q-learning, our variant of actor-critic also operates in the forward view and uses the same mix of n-step returns to update both the policy and the value-function.
The policy and the value function are updated after every $t_{max}$ actions or when a terminal state is reached.
The update performed by the algorithm can be seen as $\nabla_{\theta'} \log\pi(a_t|s_t;\theta') A(s_t,a_t;\theta,\theta_v)$ where $A(s_t,a_t;\theta,\theta_v)$ is an estimate of the advantage function given by $\sum_{i=0}^{k-1}\gamma^i r_{t+i} + \gamma^k V(s_{t+k};\theta_v)-V(s_t;\theta_v)$, where $k$ can vary from state to state and is upper-bounded by $t_{max}$.
The pseudocode for the algorithm is presented in Supplementary Algorithm~\ref{alg:reinforce}.

As with the value-based methods we rely on parallel actor-learners and accumulated updates for improving training stability.
Note that while the parameters $\theta$ of the policy and $\theta_v$ of the value function are shown as being separate for generality, we always share some of the parameters in practice.
We typically use a convolutional neural network that has one softmax output for the policy $\pi(a_t|s_t;\theta)$ and one linear output for the value function $V(s_t;\theta_v)$, with all non-output layers shared.

We also found that adding the entropy of the policy $\pi$ to the objective function improved exploration by discouraging premature convergence to suboptimal deterministic policies.
This technique was originally proposed by~\citep{williams1991function}, who found that it was particularly helpful on tasks requiring hierarchical behavior.
The gradient of the full objective function including the entropy regularization term with respect to the policy parameters takes the form
$\nabla_{\theta'} \log\pi(a_t|s_t;\theta') (R_t - V(s_t;\theta_v)) + \beta \nabla_{\theta'} H(\pi(s_t;\theta'))$,
where $H$ is the entropy.  The hyperparameter $\beta$ controls the strength of the entropy regularization term.

\textbf{Optimization:}
We investigated three different optimization algorithms in our asynchronous framework -- SGD with momentum, RMSProp~\citep{tieleman2012lecture} without shared statistics, and RMSProp with shared statistics.
We used the standard non-centered RMSProp update given by
\begin{eqnarray}
\label{eq-rmsprop}
g = \alpha g + (1-\alpha)\Delta\theta^2 \text{ and } \theta \gets \theta - \eta \frac{\Delta\theta}{\sqrt{g+\epsilon}},
\end{eqnarray}
where all operations are performed elementwise.
A comparison on a subset of Atari 2600 games showed that a variant of RMSProp where statistics $g$ are shared across threads is considerably more robust than the other two methods.
Full details of the methods and comparisons are included in Supplementary Section~\ref{sec:opt}.

\vspace{-0.1cm}
\section{Experiments}
\label{sec:experiments}
\begin{figure*}[t]
\begin{center}
\includegraphics[width=\textwidth]{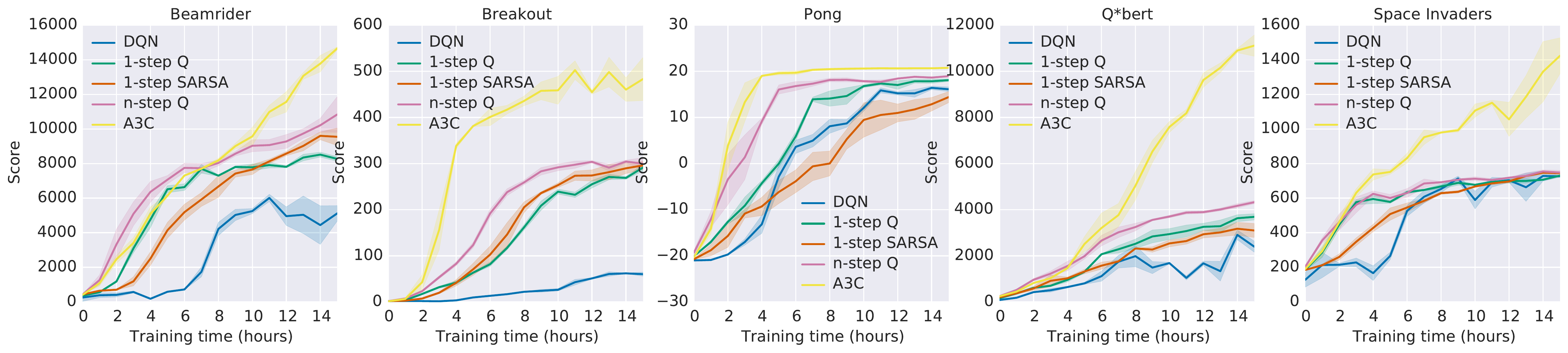}
\caption{\label{fig-speed} Learning speed comparison for DQN and the new asynchronous algorithms on five Atari 2600 games. DQN was trained on a single Nvidia K40 GPU while the asynchronous methods were trained using 16 CPU cores. The plots are averaged over 5 runs.  In the case of DQN the runs were for different seeds with fixed hyperparameters.  For asynchronous methods we average over the best 5 models from 50 experiments with learning rates sampled from $LogUniform(10^{-4},10^{-2})$ and all other hyperparameters fixed.}
\end{center}
\vspace{-0.45cm}
\end{figure*}

We use four different platforms for assessing the properties of the proposed framework.
We perform most of our experiments using the Arcade Learning Environment~\citep{bellemare-ale}, which provides a simulator for Atari 2600 games.
This is one of the most commonly used benchmark environments for RL algorithms.
We use the Atari domain to 
compare against state of the art results~\cite{hado2015doubledqn,wang2015dueling,schaul2015prioritized,nair2015gorila,mnih-dqn-2015},
as well as to carry out a detailed stability and scalability analysis of the proposed methods.
We performed further comparisons using the TORCS 3D car racing simulator~\citep{wymann-torcs}.
We also use two additional domains to evaluate only the A3C algorithm -- Mujoco and Labyrinth.  MuJoCo~\citep{todorov-mujoco} is a physics simulator for evaluating agents on continuous motor control tasks with contact dynamics.
Labyrinth is a new 3D environment where the agent must learn to find rewards in randomly generated mazes from a visual input.
The precise details of our experimental setup can be found in Supplementary Section~\ref{sec:experimental-setup}. 

\subsection{Atari 2600 Games}

We first present results on a subset of Atari 2600 games to demonstrate the training speed of the new methods.
Figure~\ref{fig-speed} compares the learning speed of the DQN algorithm trained on an Nvidia K40 GPU with the asynchronous methods trained using 16 CPU cores on five Atari 2600 games.
The results show that all four asynchronous methods we presented can successfully train neural network controllers on the Atari domain.
The asynchronous methods tend to learn faster than DQN, with significantly faster learning on some games, while training on only 16 CPU cores.
Additionally, the results suggest that n-step methods learn faster than one-step methods on some games.
Overall, the policy-based advantage actor-critic method significantly outperforms all three value-based methods.

We then evaluated asynchronous advantage actor-critic on 57 Atari games.
In order to compare with the state of the art in Atari game playing, we largely followed the training
and evaluation protocol of~\cite{hado2015doubledqn}.
Specifically, we tuned hyperparameters (learning rate and amount of gradient norm clipping) using a search on six Atari games (Beamrider, Breakout, Pong, Q*bert, Seaquest and Space Invaders) and then fixed all hyperparameters for all 57 games.
We trained both a feedforward agent with the same architecture as~\citep{mnih-dqn-2015,nair2015gorila,hado2015doubledqn} as well as a recurrent agent with an additional 256 LSTM cells after the final hidden layer.
We additionally used the final network weights for evaluation to make the results more comparable to the original results from~\cite{bellemare-ale}.
We trained our agents for four days using 16 CPU cores, while the other agents were trained for 8 to 10 days on Nvidia K40 GPUs.
Table~\ref{fig-atari-stats} shows the average and median human-normalized scores obtained by our agents trained by asynchronous advantage actor-critic (A3C) as well as
the current state-of-the art. Supplementary Table~\ref{fig-atari-full} shows the scores on all games.
A3C significantly improves on state-of-the-art the average score over 57 games
in half the training time of the other methods while using only 16 CPU cores and no GPU.
Furthermore, after just one day of training, A3C matches the average human normalized score of Dueling Double DQN and almost reaches the median human normalized score of Gorila. We note that many of the improvements that are presented in Double DQN~\citep{hado2015doubledqn} and Dueling Double DQN~\citep{wang2015dueling} can be incorporated to 1-step Q and n-step Q methods presented in this work with similar potential improvements.

\begin{table}
\small
\begin{center}
\begin{tabular}{ | l | c | c | c | }
\hline
Method & Training Time & Mean & Median \\
\hline\hline
DQN & 8 days on GPU & 121.9\% & 47.5\% \\
Gorila & 4 days, 100 machines & 215.2\% & 71.3\% \\
D-DQN & 8 days on GPU & 332.9\% & 110.9\% \\
Dueling D-DQN & 8 days on GPU & 343.8\% & 117.1\% \\
Prioritized DQN & 8 days on GPU & 463.6\% & 127.6\% \\
A3C, FF & 1 day on CPU & 344.1\% & 68.2\% \\
A3C, FF & 4 days on CPU & 496.8\% & 116.6\% \\
A3C, LSTM & 4 days on CPU & 623.0\% & 112.6\% \\

\hline
\end{tabular}
\caption{\label{fig-atari-stats}
    Mean and median human-normalized scores on 57 Atari games using the human starts evaluation metric.  Supplementary Table~S\ref{fig-atari-full} shows the raw scores for all games.
}
\end{center}
\vspace{-0.45cm}
\end{table}

\subsection{TORCS Car Racing Simulator}

We also compared the four asynchronous methods on the TORCS 3D car racing game~\citep{wymann-torcs}.
TORCS not only has more realistic graphics than Atari 2600 games, but also requires the agent to learn the dynamics of the car it is controlling.
At each step, an agent received only a visual input in the form of an RGB image of the current frame as well as a reward proportional to the agent's velocity along the center of the track at the agent's current position.
We used the same neural network architecture as the one used in the Atari experiments specified in Supplementary Section~\ref{sec:experimental-setup}.
We performed experiments using four different settings -- the agent controlling a slow car with and without opponent bots, and the agent controlling a fast car with and without opponent bots.
Full results can be found in Supplementary Figure~\ref{fig-torcs}.
A3C was the best performing agent, reaching between roughly $75\%$ and $90\%$ of the score obtained by a human tester on all four game configurations in about 12 hours of training.
A video showing the learned driving behavior of the A3C agent can be found at \url{https://youtu.be/0xo1Ldx3L5Q}.

\subsection{Continuous Action Control Using the MuJoCo Physics Simulator}
We also examined a set of tasks where the action space is continuous.
In particular, we looked at a set of rigid body physics domains with contact dynamics
where the tasks include many examples of manipulation and locomotion.
These tasks were simulated using the Mujoco physics engine.
We evaluated only the asynchronous advantage actor-critic algorithm since, unlike the value-based methods, it is easily extended to continuous actions.
In all problems, using either the physical state or pixels as input, Asynchronous Advantage-Critic found good solutions in less than 24 hours of training and typically in under a few hours.
Some successful policies learned by our agent can be seen in the following video \url{https://youtu.be/Ajjc08-iPx8}.
Further details about this experiment can be found in Supplementary Section~\ref{sec:mujoco}.

\subsection{Labyrinth}

We performed an additional set of experiments with A3C on a new 3D environment called Labyrinth.
The specific task we considered involved the agent learning to find rewards in randomly generated mazes.
At the beginning of each episode the agent was placed in a new randomly generated maze consisting of rooms and corridors.
Each maze contained two types of objects that the agent was rewarded for finding -- apples and portals.
Picking up an apple led to a reward of $1$.  Entering a portal led to a reward of $10$ after which the agent was respawned in a new random location in the maze and all previously collected apples were regenerated.
An episode terminated after 60 seconds after which a new episode would begin.
The aim of the agent is to collect as many points as possible in the time limit and the optimal strategy involves first finding the portal and then repeatedly going back to it after each respawn.
This task is much more challenging than the TORCS driving domain because the agent is faced with a new maze in each episode and must learn a general strategy for exploring random mazes.

We trained an A3C LSTM agent on this task using only $84\times 84$ RGB images as input.
The final average score of around 50 indicates that the agent learned a reasonable strategy for exploring random 3D maxes using only a visual input.
A video showing one of the agents exploring previously unseen mazes is included at \url{https://youtu.be/nMR5mjCFZCw}.

\subsection{Scalability and Data Efficiency}

\begin{table}[t]
\small
\begin{center}
\begin{tabular}{ | l | c | c | c | c | c | }
\hline
& \multicolumn{5}{|c|}{Number of threads} \\ \hline
Method & 1 & 2 & 4 & 8 & 16 \\
\hline
1-step Q & 1.0 & \textbf{3.0} & \textbf{6.3} & \textbf{13.3} & \textbf{24.1} \\ \hline
1-step SARSA & 1.0 & \textbf{2.8} & \textbf{5.9} & \textbf{13.1} & \textbf{22.1} \\ \hline
n-step Q & 1.0 & \textbf{2.7} & \textbf{5.9} & \textbf{10.7} & \textbf{17.2} \\ \hline
A3C & 1.0 & 2.1 & 3.7 & 6.9 & 12.5 \\ \hline
\end{tabular}
\caption{\label{fig-scalability} The average training speedup for each method and number of threads averaged over seven Atari games.
To compute the training speed-up on a single game we measured the time to required reach a fixed reference score using each method and number of threads.
The speedup from using $n$ threads on a game was defined as the time required to reach a fixed reference score using one thread divided the time required to reach the reference score using $n$ threads.
The table shows the speedups averaged over seven Atari games (Beamrider, Breakout, Enduro, Pong, Q*bert, Seaquest, and Space Invaders).
}
\end{center}
\vspace{-0.45cm}
\end{table}

\begin{figure*}[ht]
\centerline{\includegraphics[width=0.95\textwidth]{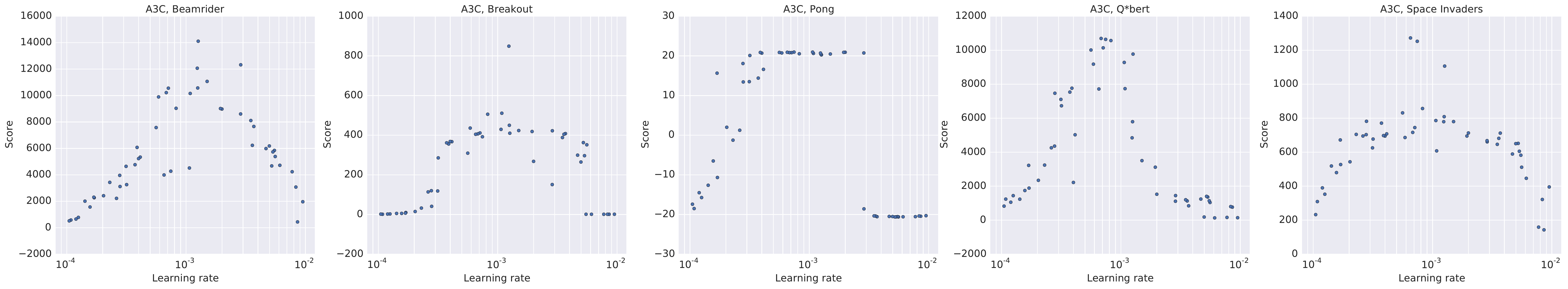}}
\vspace{-0.25cm}
\caption{\label{fig-stability-ac} Scatter plots of scores obtained by asynchronous advantage actor-critic on five games (Beamrider, Breakout, Pong, Q*bert, Space Invaders) for $50$ different learning rates and random initializations. On each game, there is a wide range of learning rates for which all random initializations acheive good scores.  This shows that A3C is quite robust to learning rates and initial random weights.}
\vspace{-0.45cm}
\end{figure*}

We analyzed the effectiveness of our proposed framework by looking at how the training time and data efficiency changes with the number of parallel actor-learners.
When using multiple workers in parallel and updating a shared model, one would expect that in an ideal case, for a given task and algorithm, the number of training steps to achieve a certain score would remain the same with varying numbers of workers.
Therefore, the advantage would be solely due to the ability of the system to consume more data in the same amount of wall clock time and possibly improved exploration.
Table~\ref{fig-scalability} shows the training speed-up achieved by using increasing numbers of parallel actor-learners averaged over seven Atari games.
These results show that all four methods achieve substantial speedups from using multiple worker threads, with $16$ threads leading to at least an order of magnitude speedup.
This confirms that our proposed framework scales well with the number of parallel workers, making efficient use of resources.

\begin{figure*}
    \centerline{\includegraphics[width=0.94\textwidth]{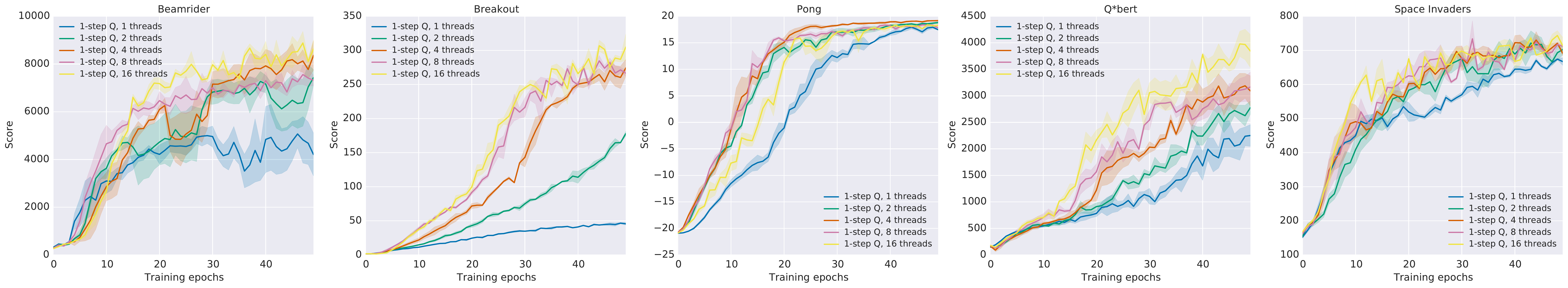}}
    \centerline{\includegraphics[width=0.94\textwidth]{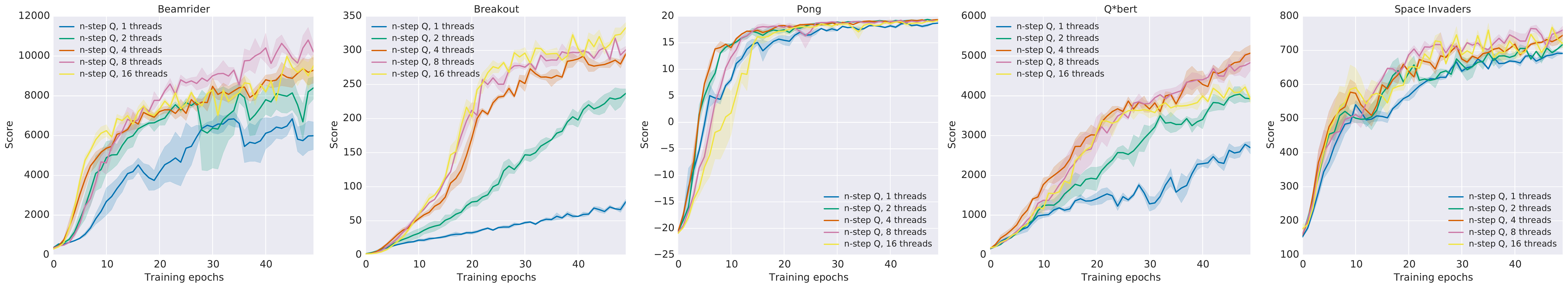}}
    \centerline{\includegraphics[width=0.94\textwidth]{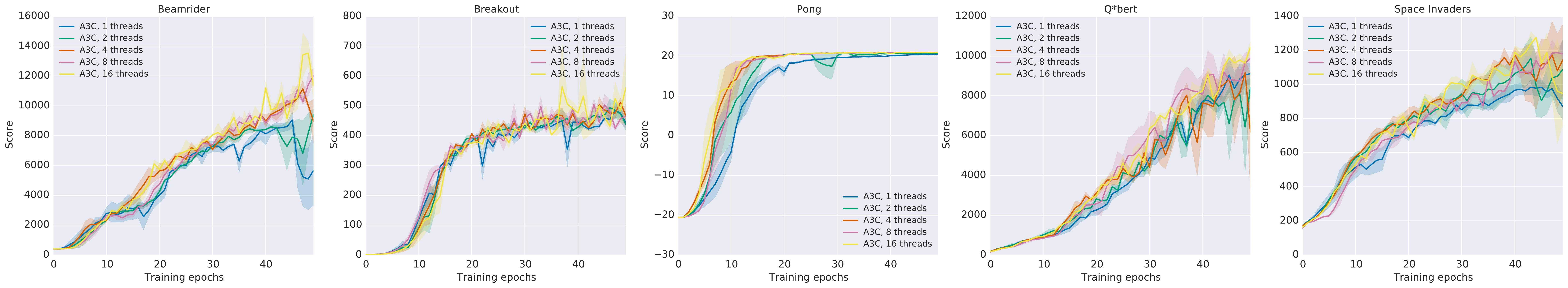}}
    \caption{\label{fig-scalability-data} Data efficiency comparison of different numbers of actor-learners for three asynchronous methods on five Atari games. The x-axis shows the total number of training epochs where an epoch corresponds to four million frames (across all threads).  The y-axis shows the average score. Each curve shows the average over the three best learning rates. Single step methods show increased data efficiency from more parallel workers. Results for Sarsa are shown in Supplementary Figure~\ref{fig-scalability-data-sarsa}.}
\vspace{-0.6cm}
\end{figure*}

\begin{figure*}
    \centerline{\includegraphics[width=0.94\textwidth]{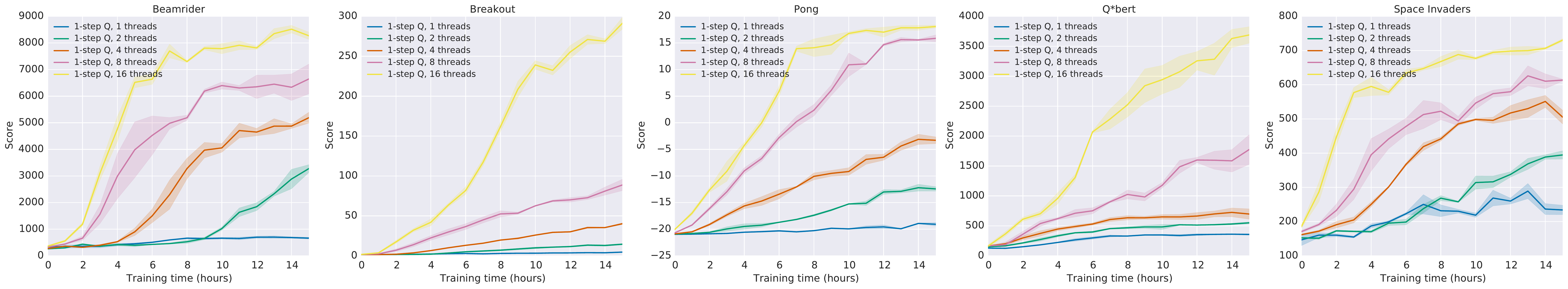}}
    \centerline{\includegraphics[width=0.94\textwidth]{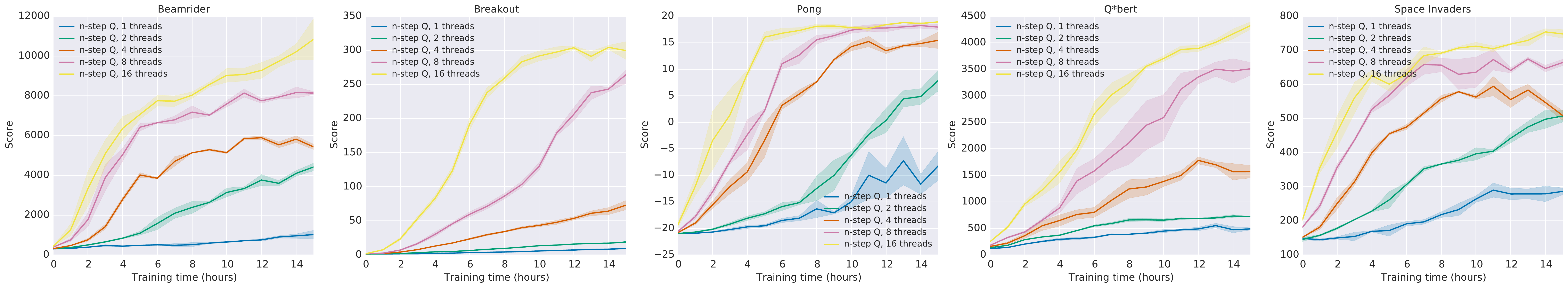}}
    \centerline{\includegraphics[width=0.94\textwidth]{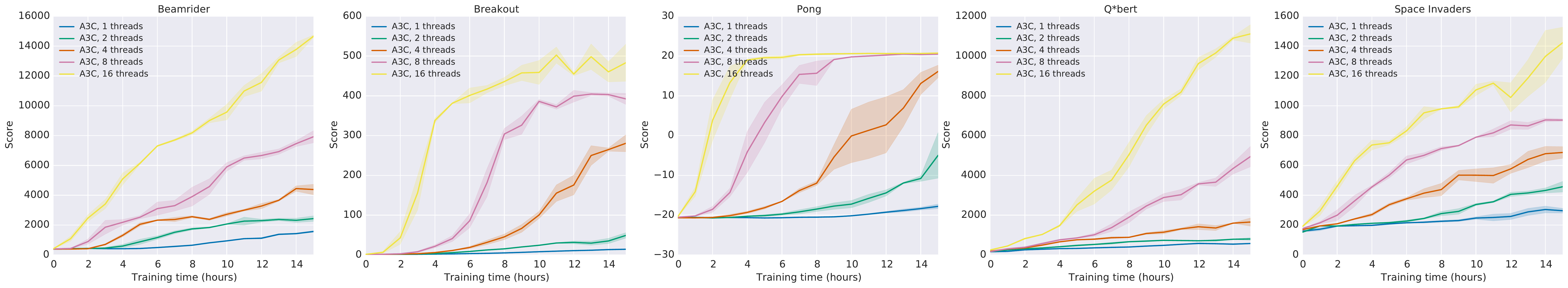}}
    \caption{\label{fig-scalability-time} Training speed comparison of different numbers of actor-learners on five Atari games. The x-axis shows training time in hours while the y-axis shows the average score.  Each curve shows the average over the three best learning rates. All asynchronous methods show significant speedups from using greater numbers of parallel actor-learners.  Results for Sarsa are shown in Supplementary Figure~\ref{fig-scalability-time-sarsa}.}
\vspace{-0.6cm}
\end{figure*}

Somewhat surprisingly, asynchronous one-step Q-learning and Sarsa algorithms exhibit superlinear speedups that cannot be explained by purely computational gains.
We observe that one-step methods (one-step Q and one-step Sarsa) often require less data to achieve a particular score when using more parallel actor-learners.
We believe this is due to positive effect of multiple threads to reduce the bias in one-step methods.
These effects are shown more clearly in Figure~\ref{fig-scalability-data}, which shows plots of the average score against the total number of training frames for different numbers of actor-learners and training methods on five Atari games, and Figure~\ref{fig-scalability-time}, which shows plots of the average score against wall-clock time.

\subsection{Robustness and Stability}
Finally, we analyzed the stability and robustness of the four proposed asynchronous algorithms.
For each of the four algorithms we trained models on five games (Breakout, Beamrider, Pong, Q*bert, Space Invaders) using $50$ different learning rates and random initializations.
Figure~\ref{fig-stability-ac} shows scatter plots of the resulting scores for A3C, while Supplementary Figure~\ref{fig-stability-lr} shows plots for the other three methods.
There is usually a range of learning rates for each method and game combination that leads to good scores, indicating that all methods are quite robust to the choice of learning rate and random initialization.
The fact that there are virtually no points with scores of $0$ in regions with good learning rates indicates that the methods are stable and do not collapse or diverge once they are learning.


\vspace{-0.1cm}
\section{Conclusions and Discussion}

We have presented asynchronous versions of four standard reinforcement learning algorithms and showed that they are able to train neural network controllers on a variety of domains in a stable manner.
Our results show that in our proposed framework stable training of neural networks through reinforcement learning is possible with both value-based and policy-based methods, off-policy as well as on-policy methods, and in discrete as well as continuous domains.
When trained on the Atari domain using 16 CPU cores, the proposed asynchronous algorithms train faster than DQN trained on an Nvidia K40 GPU, with A3C surpassing the current state-of-the-art in half the training time.

One of our main findings is that using parallel actor-learners to update a shared model had a stabilizing effect on the learning process of the three value-based methods we considered.
While this shows that stable online Q-learning is possible without experience replay, which was used for this purpose in DQN, it does not mean that experience replay is not useful.
Incorporating experience replay into the asynchronous reinforcement learning framework could substantially improve the data efficiency of these methods by reusing old data.
This could in turn lead to much faster training times in domains like TORCS where interacting with the environment is more expensive than updating the model for the architecture we used.

Combining other existing reinforcement learning methods or recent advances in deep reinforcement learning with our asynchronous framework presents many possibilities for immediate improvements to the methods we presented.
While our n-step methods operate in the \emph{forward view}~\citep{sutton:book} by using corrected n-step returns directly as targets, it has been more common to use the \emph{backward view} to implicitly combine different returns through eligibility traces~\citep{watkins1989learning,sutton:book,peng1996msq}.
The asynchronous advantage actor-critic method could be potentially improved by using other ways of estimating the advantage function, such as generalized advantage estimation of~\cite{schulman2015gae}.
All of the value-based methods we investigated could benefit from different ways of reducing over-estimation bias of Q-values~\citep{hado2015doubledqn,bellemare2016gap}.
Yet another, more speculative, direction is to try and combine the recent work on true online temporal difference methods~\citep{seijen2015true} with nonlinear function approximation.

In addition to these algorithmic improvements, a number of complementary improvements to the neural network architecture are possible.
The dueling architecture of~\cite{wang2015dueling} has been shown to produce more accurate estimates of Q-values by including separate streams for the state value and advantage in the network.
The spatial softmax proposed by~\cite{levine2015endtoend} could improve both value-based and policy-based methods by making it easier for the network to represent feature coordinates.

\subsubsection*{Acknowledgments}
We thank Thomas Degris, Remi Munos, Marc Lanctot, Sasha Vezhnevets and Joseph Modayil for many helpful discussions, suggestions and comments on the paper.
We also thank the DeepMind evaluation team for setting up the environments used to evaluate the agents in the paper.
\clearpage
\bibliographystyle{icml2016}
\bibliography{asyncrl}

\begin{thebibliography}{32}
\providecommand{\natexlab}[1]{#1}
\providecommand{\url}[1]{\texttt{#1}}
\expandafter\ifx\csname urlstyle\endcsname\relax
  \providecommand{\doi}[1]{doi: #1}\else
  \providecommand{\doi}{doi: \begingroup \urlstyle{rm}\Url}\fi

\bibitem[Bellemare et~al.(2012)Bellemare, Naddaf, Veness, and
  Bowling]{bellemare-ale}
Bellemare, Marc~G, Naddaf, Yavar, Veness, Joel, and Bowling, Michael.
\newblock The arcade learning environment: An evaluation platform for general
  agents.
\newblock \emph{Journal of Artificial Intelligence Research}, 2012.

\bibitem[Bellemare et~al.(2016)Bellemare, Ostrovski, Guez, Thomas, and
  Munos]{bellemare2016gap}
Bellemare, Marc~G., Ostrovski, Georg, Guez, Arthur, Thomas, Philip~S., and
  Munos, R{\'e}mi.
\newblock Increasing the action gap: New operators for reinforcement learning.
\newblock In \emph{Proceedings of the AAAI Conference on Artificial
  Intelligence}, 2016.

\bibitem[Bertsekas(1982)]{bertsekas1982distributed}
Bertsekas, Dimitri~P.
\newblock Distributed dynamic programming.
\newblock \emph{Automatic Control, IEEE Transactions on}, 27\penalty0
  (3):\penalty0 610--616, 1982.

\bibitem[Chavez et~al.(2015)Chavez, Ong, and Hong]{chavez2015deepq}
Chavez, Kevin, Ong, Hao~Yi, and Hong, Augustus.
\newblock Distributed deep q-learning.
\newblock Technical report, Stanford University, June 2015.

\bibitem[Degris et~al.(2012)Degris, Pilarski, and Sutton]{degris2012model}
Degris, Thomas, Pilarski, Patrick~M, and Sutton, Richard~S.
\newblock Model-free reinforcement learning with continuous action in practice.
\newblock In \emph{American Control Conference (ACC), 2012}, pp.\  2177--2182.
  IEEE, 2012.

\bibitem[Grounds \& Kudenko(2008)Grounds and Kudenko]{grounds:parallelRL}
Grounds, Matthew and Kudenko, Daniel.
\newblock Parallel reinforcement learning with linear function approximation.
\newblock In \emph{Proceedings of the 5th, 6th and 7th European Conference on
  Adaptive and Learning Agents and Multi-agent Systems: Adaptation and
  Multi-agent Learning}, pp.\  60--74. Springer-Verlag, 2008.

\bibitem[Koutn{\'\i}k et~al.(2014)Koutn{\'\i}k, Schmidhuber, and
  Gomez]{koutnik2014}
Koutn{\'\i}k, Jan, Schmidhuber, J{\"u}rgen, and Gomez, Faustino.
\newblock Evolving deep unsupervised convolutional networks for vision-based
  reinforcement learning.
\newblock In \emph{Proceedings of the 2014 conference on Genetic and
  evolutionary computation}, pp.\  541--548. ACM, 2014.

\bibitem[Levine et~al.(2015)Levine, Finn, Darrell, and
  Abbeel]{levine2015endtoend}
Levine, Sergey, Finn, Chelsea, Darrell, Trevor, and Abbeel, Pieter.
\newblock End-to-end training of deep visuomotor policies.
\newblock \emph{arXiv preprint arXiv:1504.00702}, 2015.

\bibitem[Li \& Schuurmans(2011)Li and Schuurmans]{li:mapreduce}
Li, Yuxi and Schuurmans, Dale.
\newblock Mapreduce for parallel reinforcement learning.
\newblock In \emph{Recent Advances in Reinforcement Learning - 9th European
  Workshop, {EWRL} 2011, Athens, Greece, September 9-11, 2011, Revised Selected
  Papers}, pp.\  309--320, 2011.

\bibitem[Lillicrap et~al.(2015)Lillicrap, Hunt, Pritzel, Heess, Erez, Tassa,
  Silver, and Wierstra]{lillicrap2015continuous}
Lillicrap, Timothy~P, Hunt, Jonathan~J, Pritzel, Alexander, Heess, Nicolas,
  Erez, Tom, Tassa, Yuval, Silver, David, and Wierstra, Daan.
\newblock Continuous control with deep reinforcement learning.
\newblock \emph{arXiv preprint arXiv:1509.02971}, 2015.

\bibitem[Mnih et~al.(2013)Mnih, Kavukcuoglu, Silver, Graves, Antonoglou,
  Wierstra, and Riedmiller]{mnih-atari-2013}
Mnih, Volodymyr, Kavukcuoglu, Koray, Silver, David, Graves, Alex, Antonoglou,
  Ioannis, Wierstra, Daan, and Riedmiller, Martin.
\newblock Playing atari with deep reinforcement learning.
\newblock In \emph{NIPS Deep Learning Workshop}. 2013.

\bibitem[Mnih et~al.(2015)Mnih, Kavukcuoglu, Silver, Rusu, Veness, Bellemare,
  Graves, Riedmiller, Fidjeland, Ostrovski, Petersen, Beattie, Sadik,
  Antonoglou, King, Kumaran, Wierstra, Legg, and Hassabis]{mnih-dqn-2015}
Mnih, Volodymyr, Kavukcuoglu, Koray, Silver, David, Rusu, Andrei~A., Veness,
  Joel, Bellemare, Marc~G., Graves, Alex, Riedmiller, Martin, Fidjeland,
  Andreas~K., Ostrovski, Georg, Petersen, Stig, Beattie, Charles, Sadik, Amir,
  Antonoglou, Ioannis, King, Helen, Kumaran, Dharshan, Wierstra, Daan, Legg,
  Shane, and Hassabis, Demis.
\newblock Human-level control through deep reinforcement learning.
\newblock \emph{Nature}, 518\penalty0 (7540):\penalty0 529--533, 02 2015.
\newblock URL \url{http://dx.doi.org/10.1038/nature14236}.

\bibitem[Nair et~al.(2015)Nair, Srinivasan, Blackwell, Alcicek, Fearon, Maria,
  Panneershelvam, Suleyman, Beattie, Petersen, Legg, Mnih, Kavukcuoglu, and
  Silver]{nair2015gorila}
Nair, Arun, Srinivasan, Praveen, Blackwell, Sam, Alcicek, Cagdas, Fearon, Rory,
  Maria, Alessandro~De, Panneershelvam, Vedavyas, Suleyman, Mustafa, Beattie,
  Charles, Petersen, Stig, Legg, Shane, Mnih, Volodymyr, Kavukcuoglu, Koray,
  and Silver, David.
\newblock Massively parallel methods for deep reinforcement learning.
\newblock In \emph{ICML Deep Learning Workshop}. 2015.

\bibitem[Peng \& Williams(1996)Peng and Williams]{peng1996msq}
Peng, Jing and Williams, Ronald~J.
\newblock Incremental multi-step q-learning.
\newblock \emph{Machine Learning}, 22\penalty0 (1-3):\penalty0 283--290, 1996.

\bibitem[Recht et~al.(2011)Recht, Re, Wright, and Niu]{recht2011hogwild}
Recht, Benjamin, Re, Christopher, Wright, Stephen, and Niu, Feng.
\newblock Hogwild: A lock-free approach to parallelizing stochastic gradient
  descent.
\newblock In \emph{Advances in Neural Information Processing Systems}, pp.\
  693--701, 2011.

\bibitem[Riedmiller(2005)]{riedmiller2005nfq}
Riedmiller, Martin.
\newblock Neural fitted q iteration--first experiences with a data efficient
  neural reinforcement learning method.
\newblock In \emph{Machine Learning: ECML 2005}, pp.\  317--328. Springer
  Berlin Heidelberg, 2005.

\bibitem[Rummery \& Niranjan(1994)Rummery and Niranjan]{rummery1994sarsa}
Rummery, Gavin~A and Niranjan, Mahesan.
\newblock On-line q-learning using connectionist systems.
\newblock 1994.

\bibitem[Schaul et~al.(2015)Schaul, Quan, Antonoglou, and
  Silver]{schaul2015prioritized}
Schaul, Tom, Quan, John, Antonoglou, Ioannis, and Silver, David.
\newblock Prioritized experience replay.
\newblock \emph{arXiv preprint arXiv:1511.05952}, 2015.

\bibitem[Schulman et~al.(2015{\natexlab{a}})Schulman, Levine, Moritz, Jordan,
  and Abbeel]{schulman2015trust}
Schulman, John, Levine, Sergey, Moritz, Philipp, Jordan, Michael~I, and Abbeel,
  Pieter.
\newblock Trust region policy optimization.
\newblock In \emph{International Conference on Machine Learning (ICML)},
  2015{\natexlab{a}}.

\bibitem[Schulman et~al.(2015{\natexlab{b}})Schulman, Moritz, Levine, Jordan,
  and Abbeel]{schulman2015gae}
Schulman, John, Moritz, Philipp, Levine, Sergey, Jordan, Michael, and Abbeel,
  Pieter.
\newblock High-dimensional continuous control using generalized advantage
  estimation.
\newblock \emph{arXiv preprint arXiv:1506.02438}, 2015{\natexlab{b}}.

\bibitem[Sutton \& Barto(1998)Sutton and Barto]{sutton:book}
Sutton, R. and Barto, A.
\newblock \emph{Reinforcement Learning: an Introduction}.
\newblock MIT Press, 1998.

\bibitem[Tieleman \& Hinton(2012)Tieleman and Hinton]{tieleman2012lecture}
Tieleman, Tijmen and Hinton, Geoffrey.
\newblock Lecture 6.5-rmsprop: Divide the gradient by a running average of its
  recent magnitude.
\newblock \emph{COURSERA: Neural Networks for Machine Learning}, 4, 2012.

\bibitem[Todorov(2015)]{todorov-mujoco}
Todorov, E.
\newblock \emph{MuJoCo: Modeling, Simulation and Visualization of Multi-Joint
  Dynamics with Contact (ed 1.0)}.
\newblock Roboti Publishing, 2015.

\bibitem[Tomassini(1999)]{tomassini1999parallel}
Tomassini, Marco.
\newblock Parallel and distributed evolutionary algorithms: A review.
\newblock Technical report, 1999.

\bibitem[Tsitsiklis(1994)]{tsitsiklis1994asynchronous}
Tsitsiklis, John~N.
\newblock Asynchronous stochastic approximation and q-learning.
\newblock \emph{Machine Learning}, 16\penalty0 (3):\penalty0 185--202, 1994.

\bibitem[Van~Hasselt et~al.(2015)Van~Hasselt, Guez, and
  Silver]{hado2015doubledqn}
Van~Hasselt, Hado, Guez, Arthur, and Silver, David.
\newblock Deep reinforcement learning with double q-learning.
\newblock \emph{arXiv preprint arXiv:1509.06461}, 2015.

\bibitem[{van Seijen} et~al.(2015){van Seijen}, {Rupam Mahmood}, {Pilarski},
  {Machado}, and {Sutton}]{seijen2015true}
{van Seijen}, H., {Rupam Mahmood}, A., {Pilarski}, P.~M., {Machado}, M.~C., and
  {Sutton}, R.~S.
\newblock {True Online Temporal-Difference Learning}.
\newblock \emph{ArXiv e-prints}, December 2015.

\bibitem[{Wang} et~al.(2015){Wang}, {de Freitas}, and
  {Lanctot}]{wang2015dueling}
{Wang}, Z., {de Freitas}, N., and {Lanctot}, M.
\newblock {Dueling Network Architectures for Deep Reinforcement Learning}.
\newblock \emph{ArXiv e-prints}, November 2015.

\bibitem[Watkins(1989)]{watkins1989learning}
Watkins, Christopher John Cornish~Hellaby.
\newblock \emph{Learning from delayed rewards}.
\newblock PhD thesis, University of Cambridge England, 1989.

\bibitem[Williams(1992)]{Williams1992}
Williams, R.J.
\newblock {Simple statistical gradient-following algorithms for connectionist
  reinforcement learning}.
\newblock \emph{Machine Learning}, 8\penalty0 (3):\penalty0 229--256, 1992.

\bibitem[Williams \& Peng(1991)Williams and Peng]{williams1991function}
Williams, Ronald~J and Peng, Jing.
\newblock Function optimization using connectionist reinforcement learning
  algorithms.
\newblock \emph{Connection Science}, 3\penalty0 (3):\penalty0 241--268, 1991.

\bibitem[Wymann et~al.(2013)Wymann, Espié, Guionneau, Dimitrakakis, Coulom,
  and Sumner]{wymann-torcs}
Wymann, B., Espié, E., Guionneau, C., Dimitrakakis, C., Coulom, R., and
  Sumner, A.
\newblock Torcs: The open racing car simulator, v1.3.5, 2013.

\end{thebibliography}

\clearpage

\renewcommand{\thealgorithm}{S\arabic{algorithm}}
\renewcommand{\thefigure}{S\arabic{figure}}
\renewcommand{\theequation}{S\arabic{equation}}
\renewcommand{\thetable}{S\arabic{table}}

\oddsidemargin .25in    %
\evensidemargin .25in
\marginparwidth 0.07 true in
\topmargin -0.5in
\addtolength{\headsep}{0.25in}
\textheight 8.5 true in       %
\textwidth 6.0 true in        %
\onecolumn
\title{Supplementary Material for "Asynchronous Methods for Deep Reinforcement Learning"}
\maketitle
\thispagestyle{empty}

\label{supplementary}

\section{Optimization Details}
\label{sec:opt}
We investigated two different optimization algorithms with our asynchronous framework -- stochastic gradient descent and RMSProp.  Our implementations of these algorithms do not use any locking in order to maximize throughput when using a large number of threads.

\textbf{Momentum SGD:}
The implementation of SGD in an asynchronous setting is relatively straightforward and well studied~\citep{recht2011hogwild}.
Let $\theta$ be the parameter vector that is shared across all threads and let $\Delta\theta_i$ be the accumulated gradients of the loss with respect to parameters $\theta$ computed by thread number $i$.
Each thread $i$ independently applies the standard momentum SGD update $m_i = \alpha m_i + (1-\alpha) \Delta\theta_i$ followed by $\theta \gets \theta - \eta m_i$ with learning rate $\eta$, momentum $\alpha$ and without any locks.
Note that in this setting, each thread maintains its own separate gradient and momentum vector.

\textbf{RMSProp:} While RMSProp~\citep{tieleman2012lecture} has been widely used in the deep learning literature, it has not been extensively studied in the asynchronous optimization setting.  The standard non-centered RMSProp update is given by
\begin{eqnarray}
    \label{eq-rmsprop-s}
    g = \alpha g + (1-\alpha)\Delta\theta^2 \\
    \theta \gets \theta - \eta \frac{\Delta\theta}{\sqrt{g+\epsilon}},
\end{eqnarray}
where all operations are performed elementwise.
In order to apply RMSProp in the asynchronous optimization setting one must decide whether the moving average of elementwise squared gradients $g$ is shared or per-thread.
We experimented with two versions of the algorithm.
In one version, which we refer to as RMSProp, each thread maintains its own $g$ shown in Equation~\ref{eq-rmsprop-s}.
In the other version, which we call Shared RMSProp, the vector $g$ is shared among threads and is updated asynchronously and without locking.
Sharing statistics among threads also reduces memory requirements by using one fewer copy of the parameter vector per thread.

\begin{figure}[t]
\begin{center}
\centerline{\includegraphics[width=0.8\columnwidth]{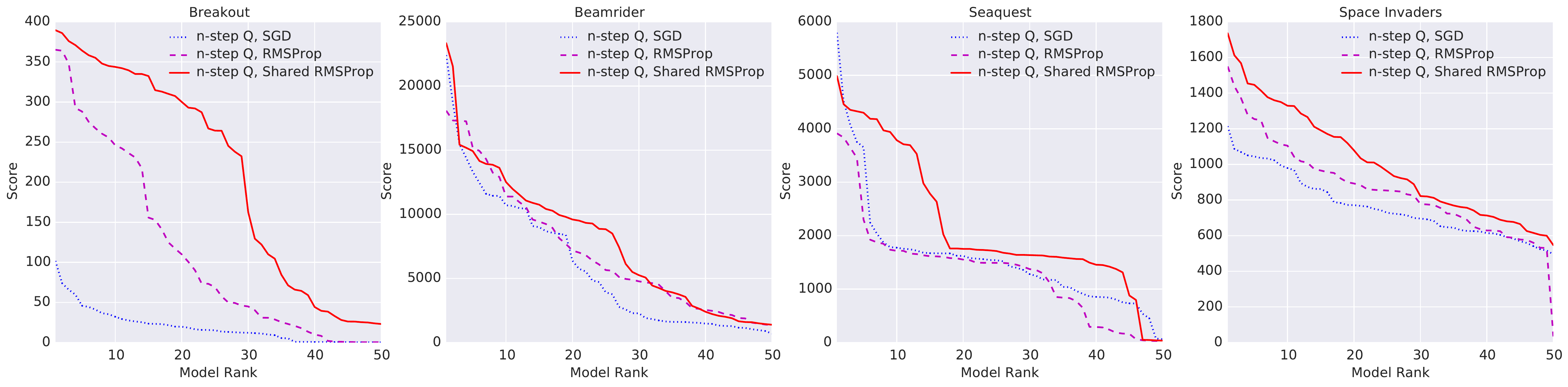}}
\centerline{\includegraphics[width=0.8\columnwidth]{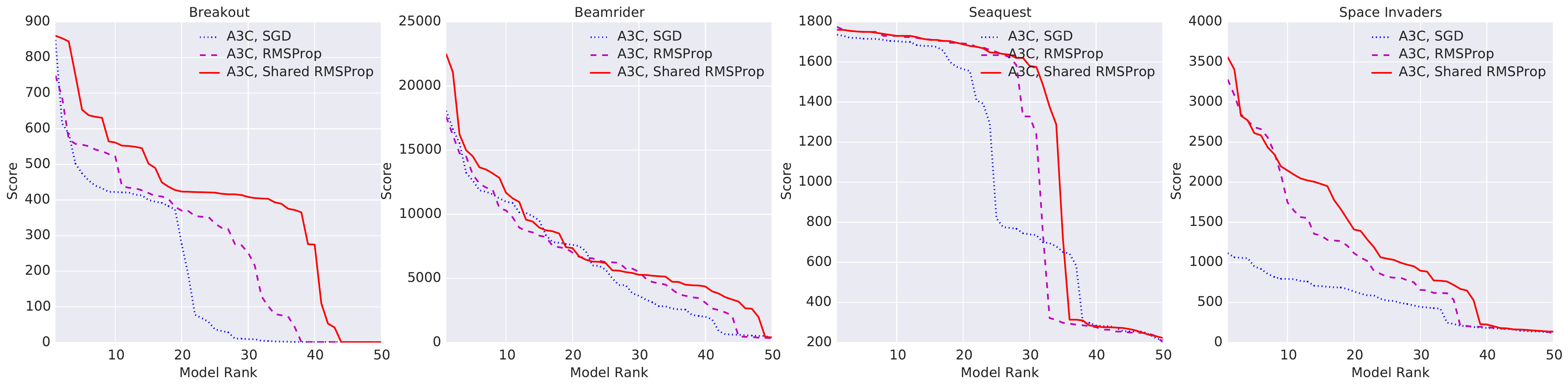}}
\caption{\label{fig-optimization} Comparison of three different optimization methods (Momentum SGD, RMSProp, Shared RMSProp) tested using two different algorithms (Async $n$-step Q and Async Advantage Actor-Critic) on four different Atari games (Breakout, Beamrider, Seaquest and Space Invaders). Each curve shows the final scores for 50 experiments sorted in descending order that covers a search over 50 random initializations and learning rates. The top row shows results using Async $n$-step Q algorithm and bottom row shows results with Async Advantage Actor-Critic. Each individual graph shows results for one of the four games and three different optimization methods. Shared RMSProp tends to be more robust to different learning rates and random initializations than Momentum SGD and RMSProp without sharing.}
\vspace{-0.5cm}
\end{center}
\end{figure}

We compared these three asynchronous optimization algorithms in terms of their sensitivity to different learning rates and random network initializations.
Figure~\ref{fig-optimization} shows a comparison of the methods for two different reinforcement learning methods (Async $n$-step Q and Async Advantage Actor-Critic) on four different games (Breakout, Beamrider, Seaquest and Space Invaders).
Each curve shows the scores for 50 experiments that correspond to 50 different random learning rates and initializations.
The x-axis shows the rank of the model after sorting in descending order by final average score and the y-axis shows the final average score achieved by the corresponding model.
In this representation, the algorithm that performs better would achieve higher maximum rewards on the y-axis and the algorithm that is most robust would have its slope closest to horizontal, thus maximizing the area under the curve.
RMSProp with shared statistics tends to be more robust than RMSProp with per-thread statistics, which is in turn more robust than Momentum SGD.

\section{Experimental Setup}
\label{sec:experimental-setup}
\begin{figure}[h]
\begin{center}
\includegraphics[width=0.4\columnwidth]{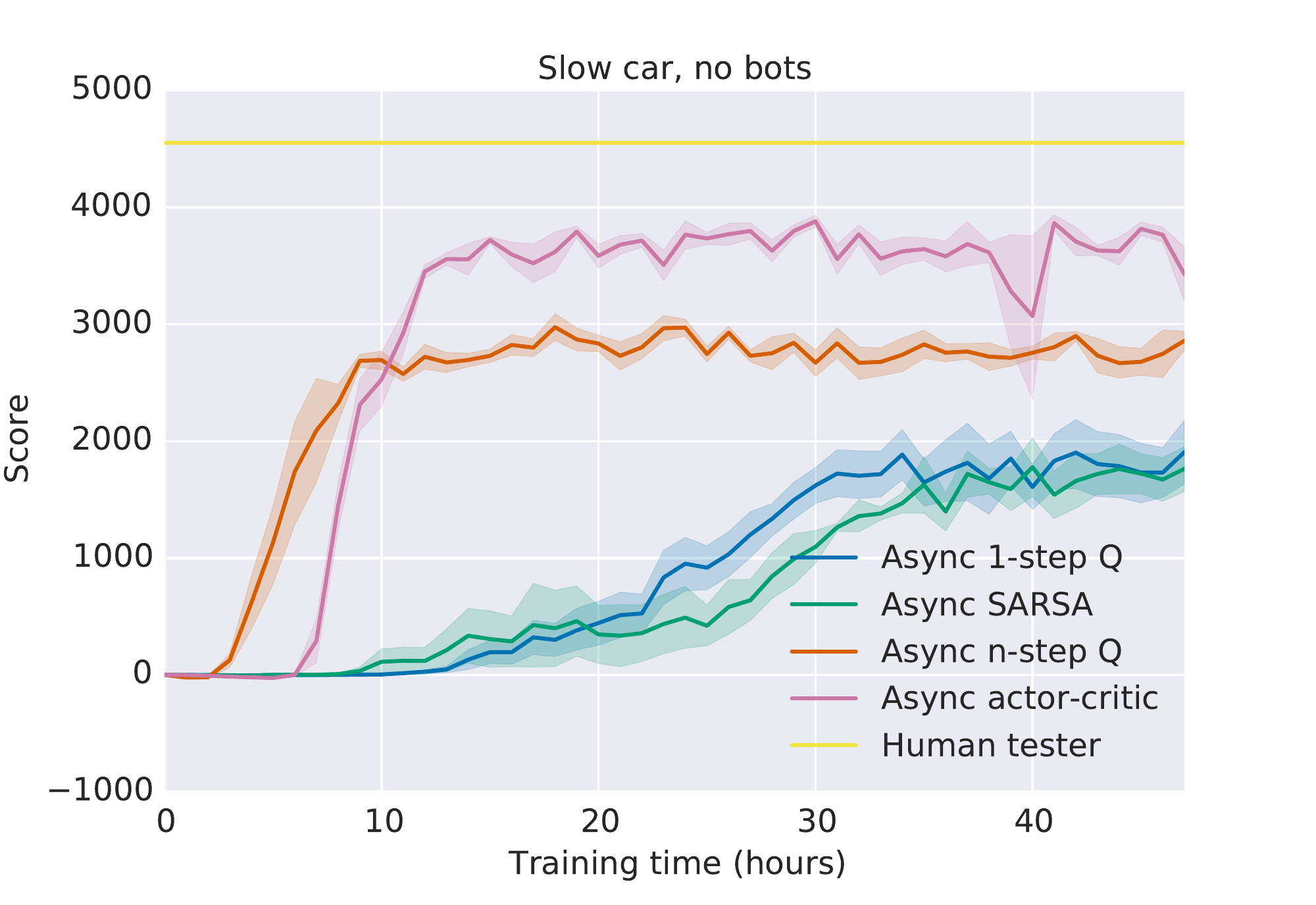}
\includegraphics[width=0.4\columnwidth]{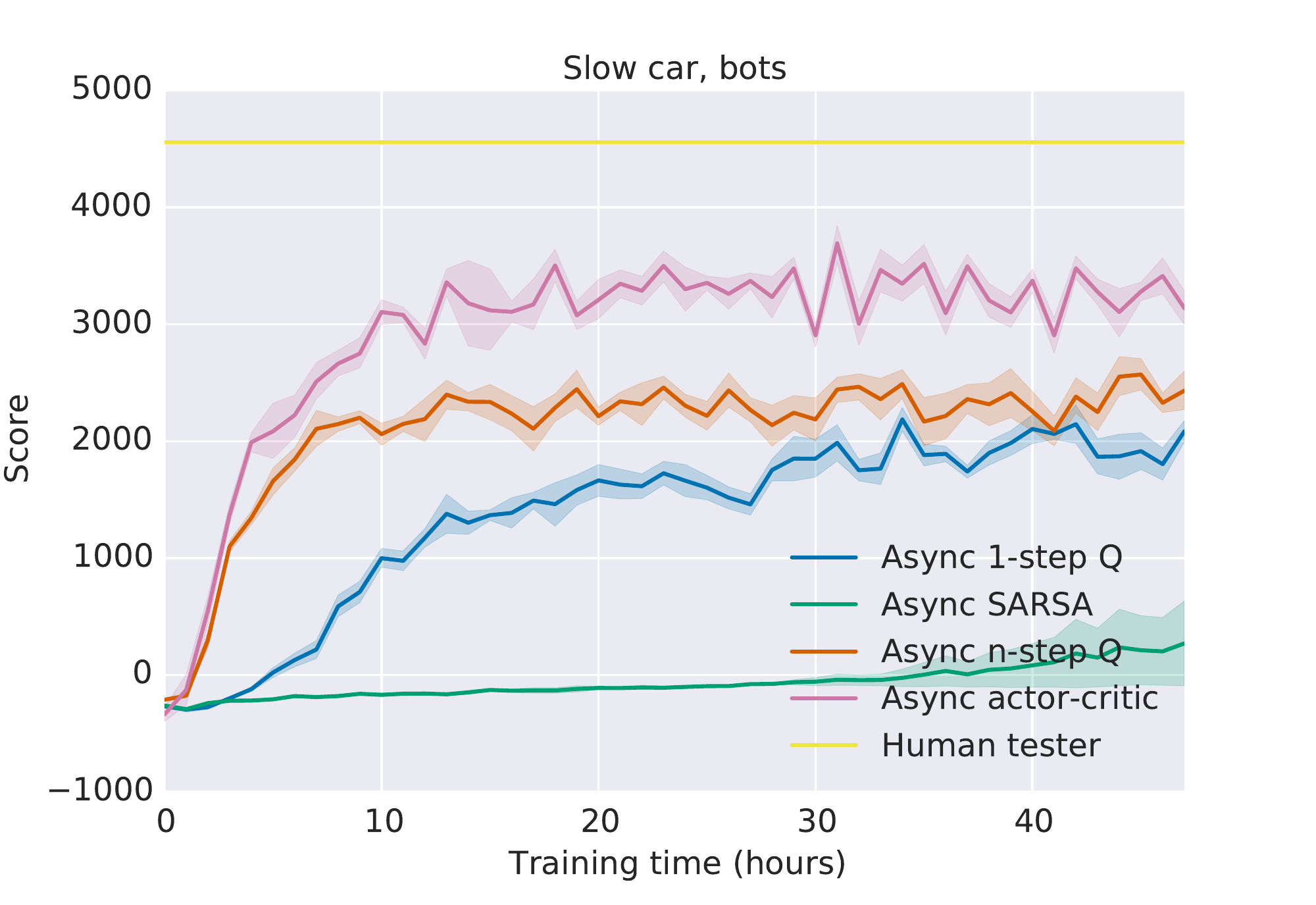} \\
\includegraphics[width=0.4\columnwidth]{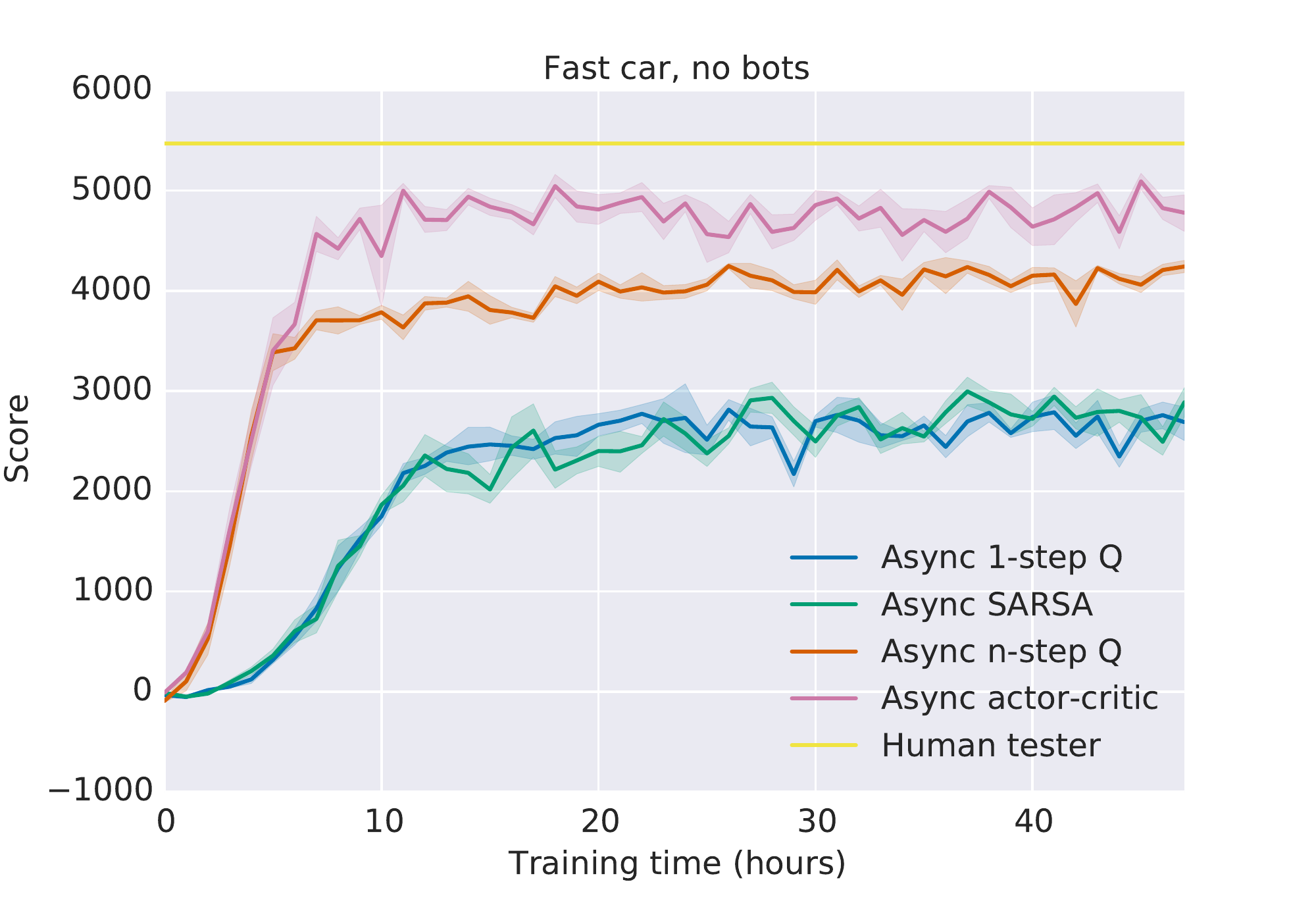}
\includegraphics[width=0.4\columnwidth]{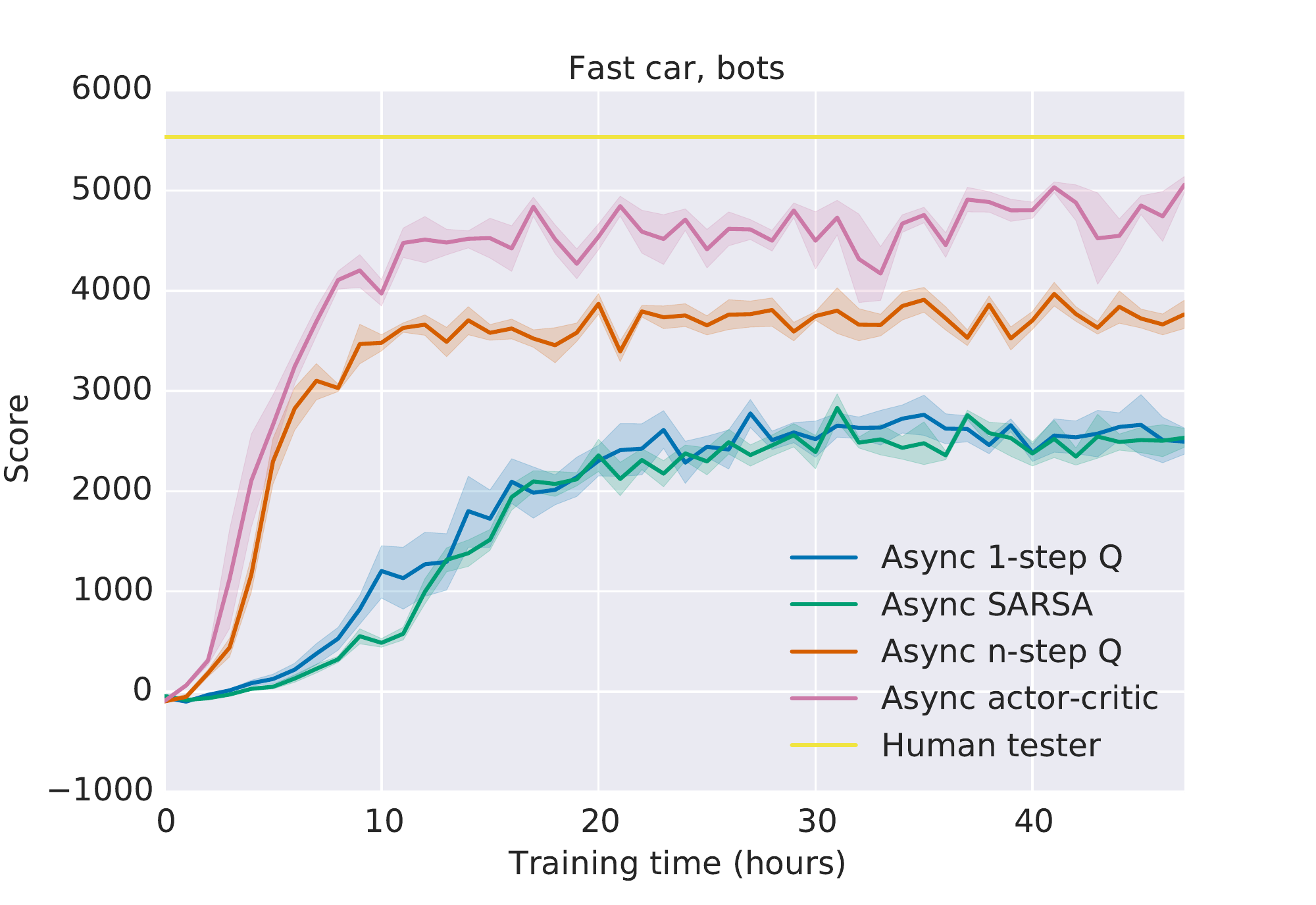}
\caption{\label{fig-torcs} Comparison of algorithms on the TORCS car racing simulator. Four different configurations of car speed and opponent presence or absence are shown. In each plot, all four algorithms (one-step Q, one-step Sarsa, $n$-step Q and Advantage Actor-Critic) are compared on score vs training time in wall clock hours. Multi-step algorithms achieve better policies much faster than one-step algorithms on all four levels. The curves show averages over the 5 best runs from 50 experiments with learning rates sampled from $LogUniform(10^{-4},10^{-2})$ and all other hyperparameters fixed.}
\vspace{-0.5cm}
\end{center}
\end{figure}

The experiments performed on a subset of Atari games (Figures~\ref{fig-speed}, \ref{fig-scalability-data}, \ref{fig-scalability-time} and Table~\ref{fig-scalability}) as well as the TORCS experiments (Figure~\ref{fig-torcs}) used the following setup.  Each experiment used 16 actor-learner threads running on a single machine and no GPUs.
All methods performed updates after every 5 actions ($t_{max}=5$ and $I_{Update}=5$) and shared RMSProp was used for optimization.
The three asynchronous value-based methods used a shared target network that was updated every 40000 frames.
The Atari experiments used the same input preprocessing as~\cite{mnih-dqn-2015} and an action repeat of 4.
The agents used the network architecture from~\cite{mnih-atari-2013}.
The network used a convolutional layer with 16 filters of size $8\times 8$ with stride 4, followed by a convolutional layer with with 32 filters of size $4\times 4$ with stride 2, followed by a fully connected layer with 256 hidden units.
All three hidden layers were followed by a rectifier nonlinearity.
The value-based methods had a single linear output unit for each action representing the action-value.
The model used by actor-critic agents had two set of outputs -- a softmax output with one entry per action representing the probability of selecting the action, and a single linear output representing the value function.
All experiments used a discount of $\gamma=0.99$ and an RMSProp decay factor of $\alpha=0.99$.

The value based methods sampled the exploration rate $\epsilon$ from a distribution taking three values $\epsilon_1, \epsilon_2, \epsilon_3$ with probabilities $0.4, 0.3, 0.3$.
The values of $\epsilon_1, \epsilon_2, \epsilon_3$ were annealed from $1$ to $0.1, 0.01, 0.5$ respectively over the first four million frames.
Advantage actor-critic used entropy regularization with a weight $\beta=0.01$ for all Atari and TORCS experiments.
We performed a set of $50$ experiments for five Atari games and every TORCS level, each using a different random initialization and initial learning rate.
The initial learning rate was sampled from a $LogUniform(10^{-4}, 10^{-2})$ distribution and annealed to $0$ over the course of training.
Note that in comparisons to prior work (Tables~\ref{fig-atari-stats} and \ref{fig-atari-full}) we followed standard evaluation protocol and used fixed hyperparameters.

\section{Continuous Action Control Using the MuJoCo Physics Simulator}
\label{sec:mujoco}

To apply the asynchronous advantage actor-critic algorithm to the Mujoco tasks the
necessary setup is nearly identical to that used in the discrete action domains, so
here we enumerate only the differences required for the continuous action domains.
The essential elements for many of the tasks (i.e. the physics models and task objectives)
are near identical to the tasks examined in \citep{lillicrap2015continuous}.
However, the rewards and thus performance are not comparable for most of the tasks due to
changes made by the developers of Mujoco which altered the contact model.

For all the domains we attempted to learn the task using the physical state as input.
The physical state consisted of the joint positions and velocities as well
as the target position if the task required a target.
In addition, for three of the tasks (pendulum, pointmass2D, and gripper) we also examined
training directly from RGB pixel inputs.
In the low dimensional physical state case, the inputs are mapped to a hidden state using one hidden layer with 200 ReLU units.
In the cases where we used pixels, the input was passed through two layers of spatial convolutions without any non-linearity or pooling.
In either case, the output of the encoder layers were fed to a single layer of 128 LSTM cells.
The most important difference in the architecture is in the the output layer of the policy
network. Unlike the discrete action domain where the action output is a Softmax, here the two outputs
of the policy network are two real number vectors which we treat
as the mean vector $\mu$ and scalar variance $\sigma^2$ of a multidimensional normal distribution
with a spherical covariance.
To act, the input is passed through the model to the output layer where we sample from the
normal distribution determined by $\mu$ and $\sigma^2$.
In practice, $\mu$ is modeled by a linear layer and $\sigma^2$ by a SoftPlus operation, $\log (1 + \exp (x))$,
as the activation computed as a function of the output of a linear layer.
In our experiments with continuous control problems the networks for policy network
and value network do not share any parameters, though this detail is unlikely to be crucial.
Finally, since the episodes were typically at most several hundred time steps long, we
did not use any bootstrapping in the policy or value function updates and batched each episode
into a single update.

\begin{figure}[t]
    \includegraphics[width=0.24\columnwidth]{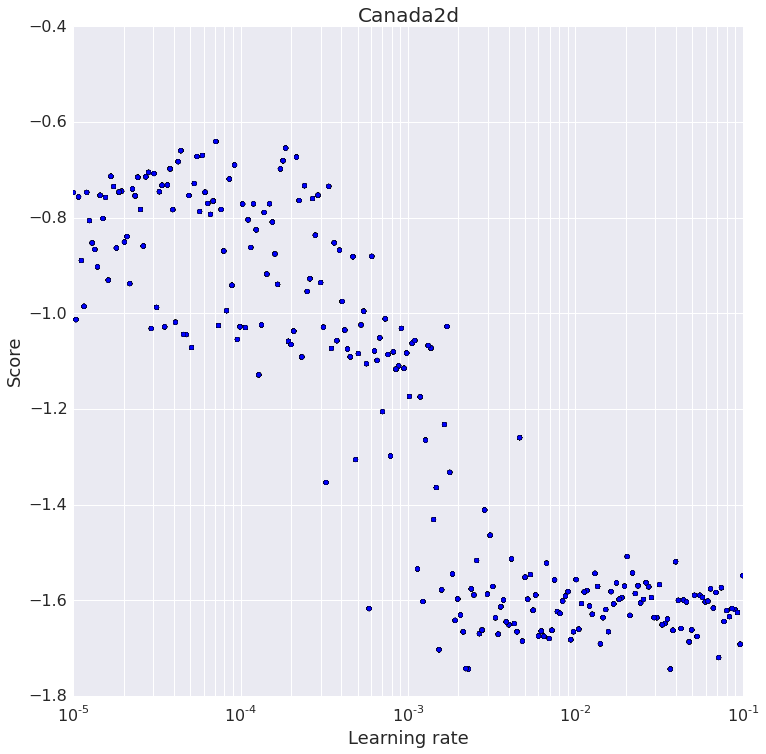}
    \includegraphics[width=0.24\columnwidth]{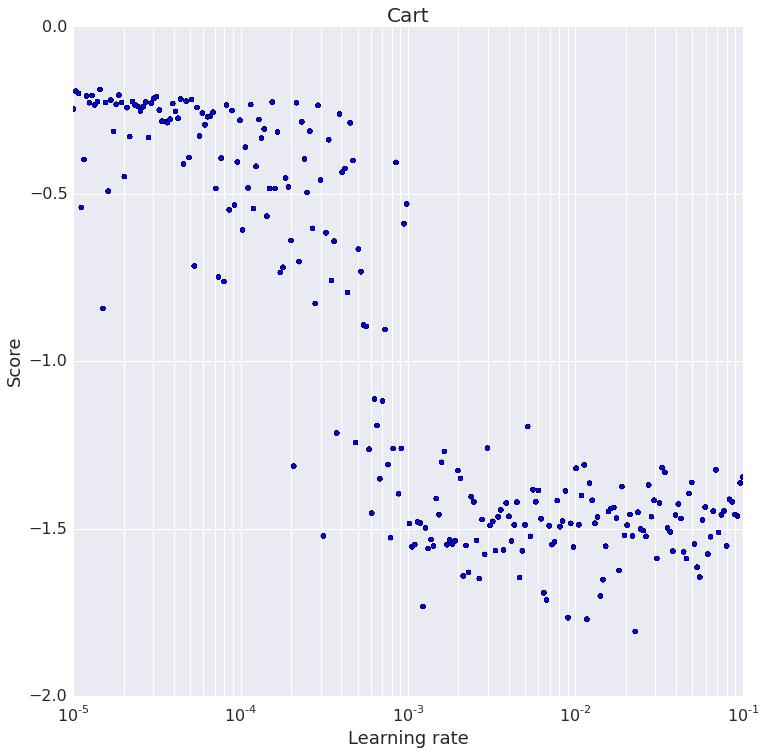}
    \includegraphics[width=0.24\columnwidth]{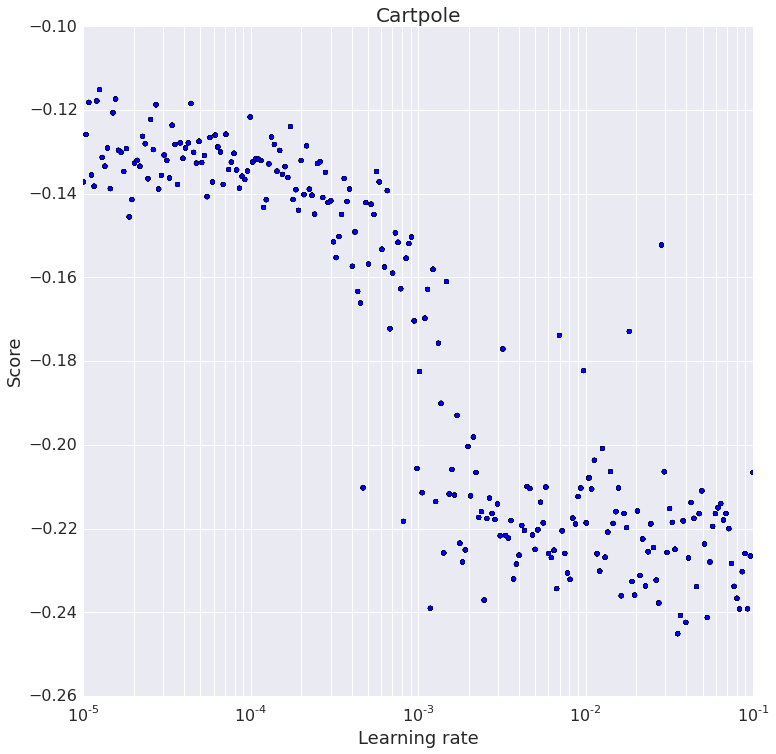}
    \includegraphics[width=0.24\columnwidth]{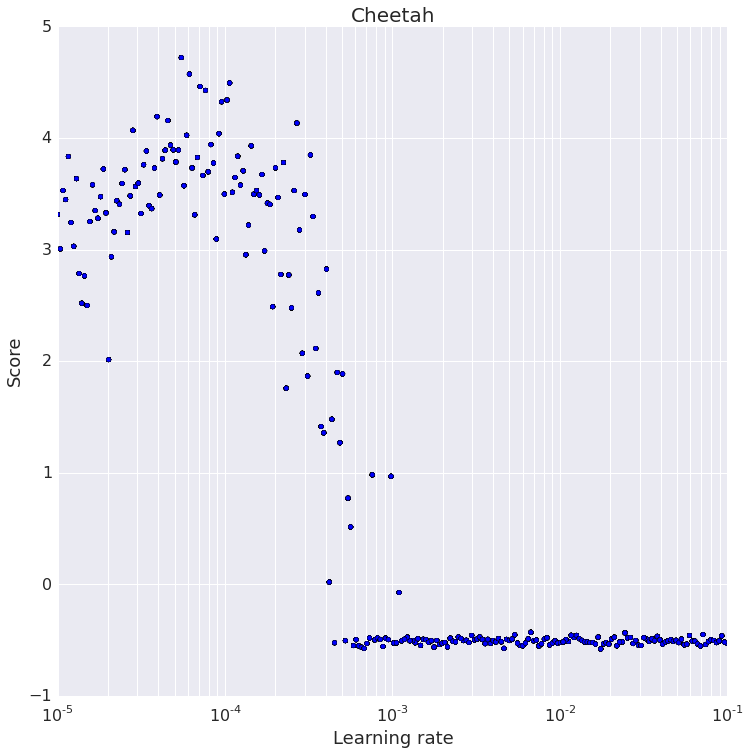}\\
    \includegraphics[width=0.24\columnwidth]{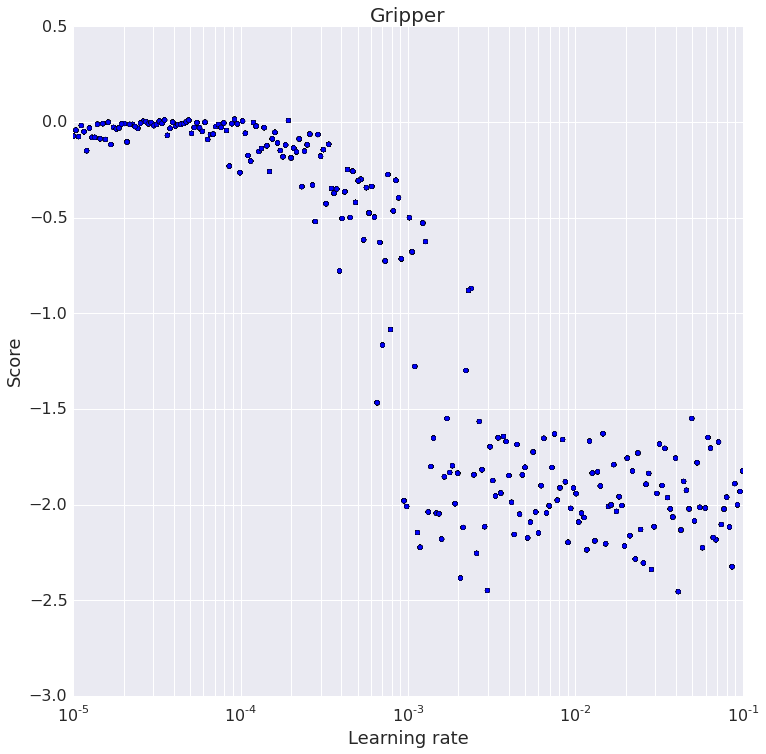}
    \includegraphics[width=0.24\columnwidth]{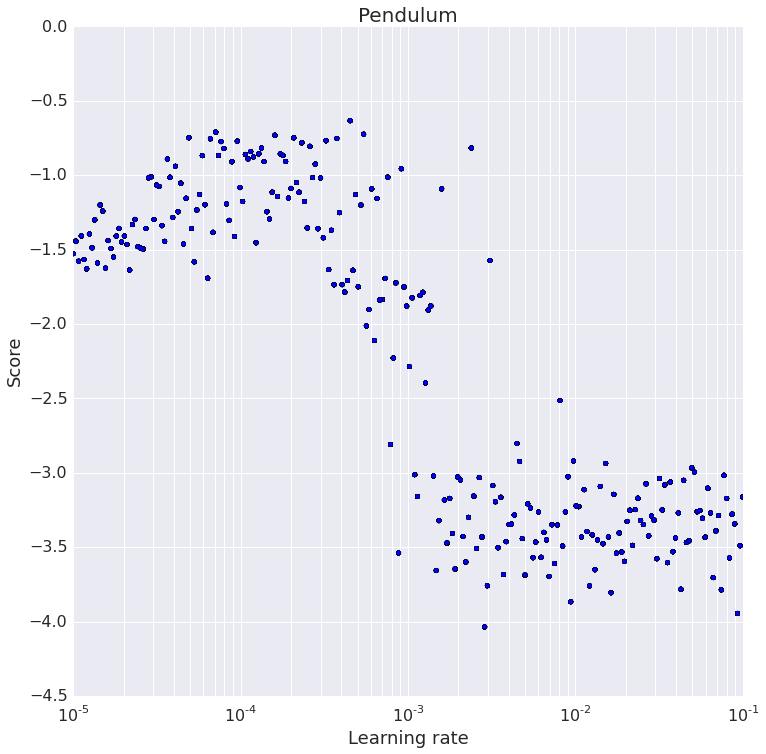}
    \includegraphics[width=0.24\columnwidth]{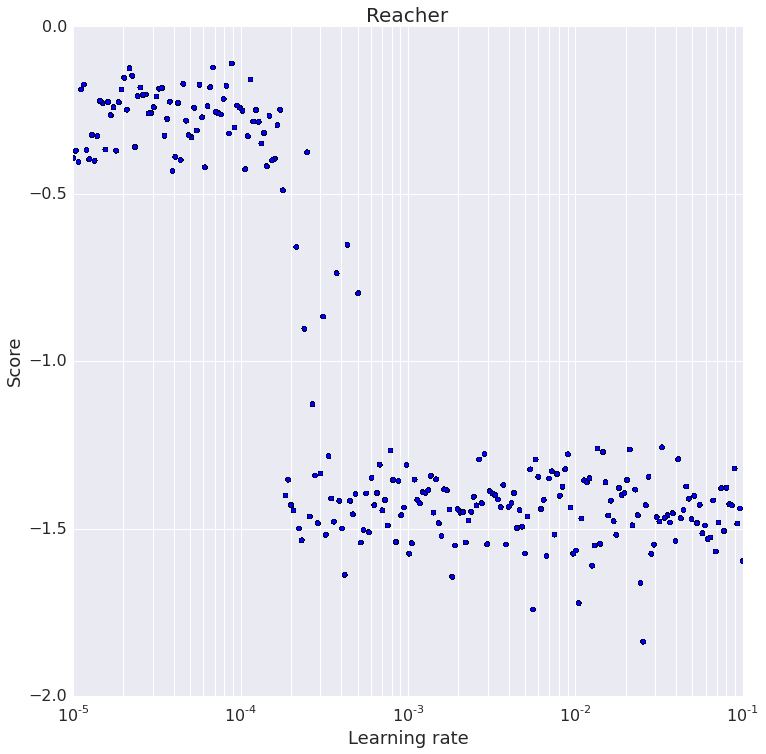}
    \includegraphics[width=0.24\columnwidth]{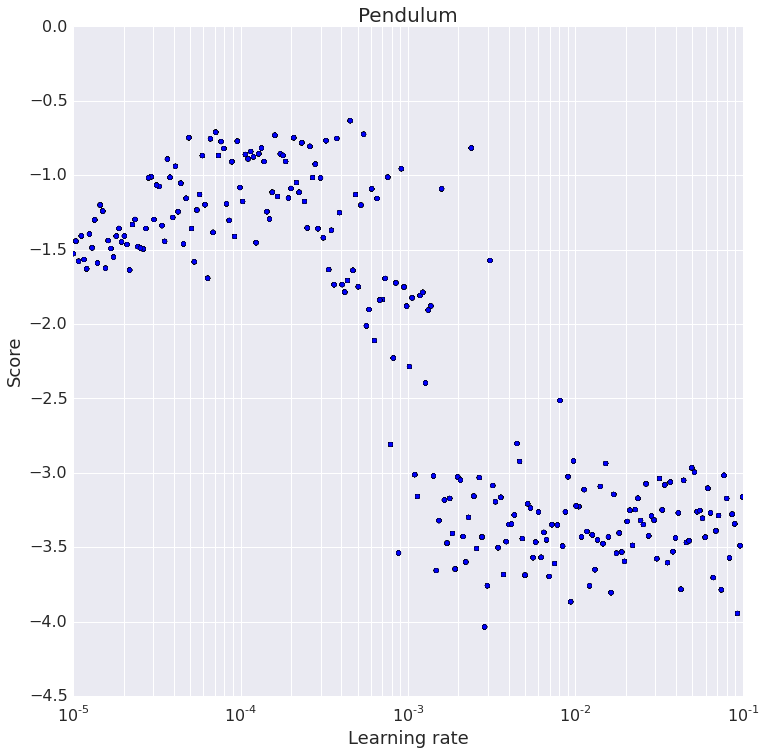}\\
    \includegraphics[width=0.24\columnwidth]{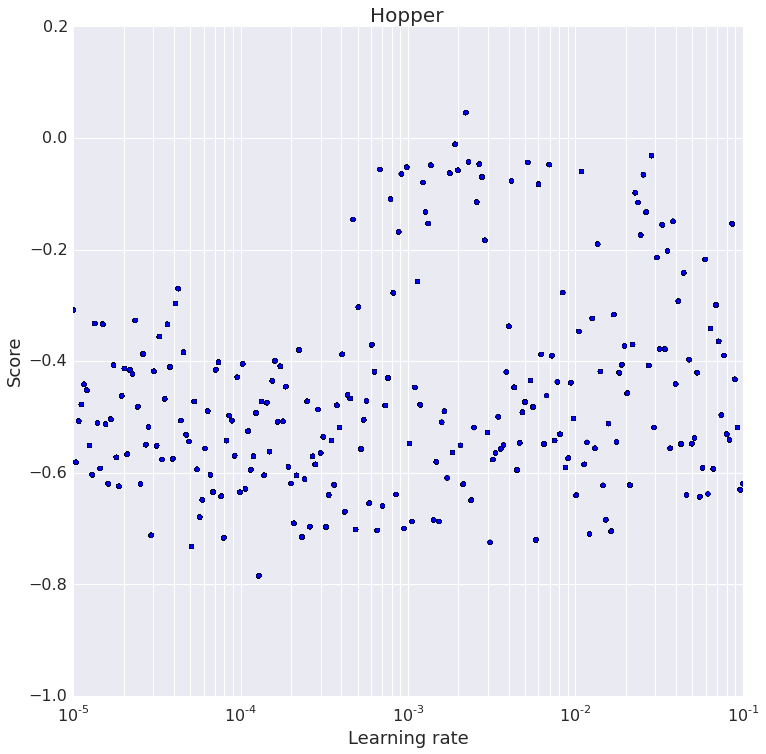}
    \includegraphics[width=0.24\columnwidth]{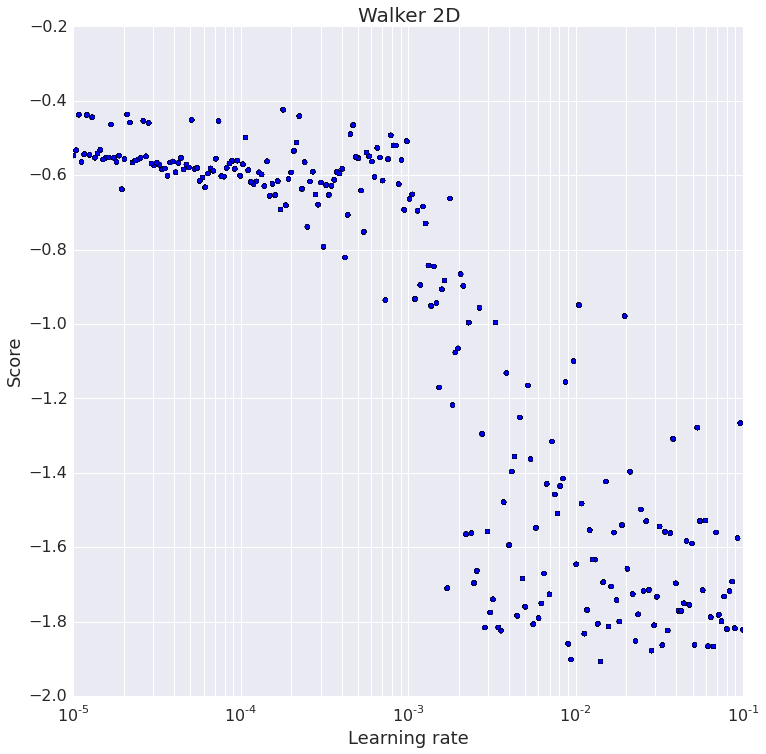}
    \includegraphics[width=0.24\columnwidth]{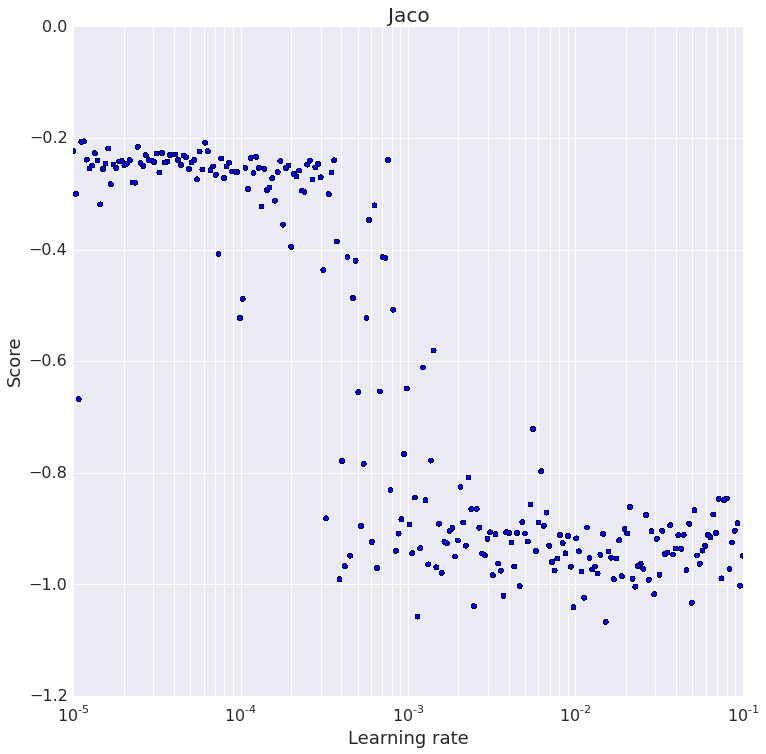}
    \includegraphics[width=0.24\columnwidth]{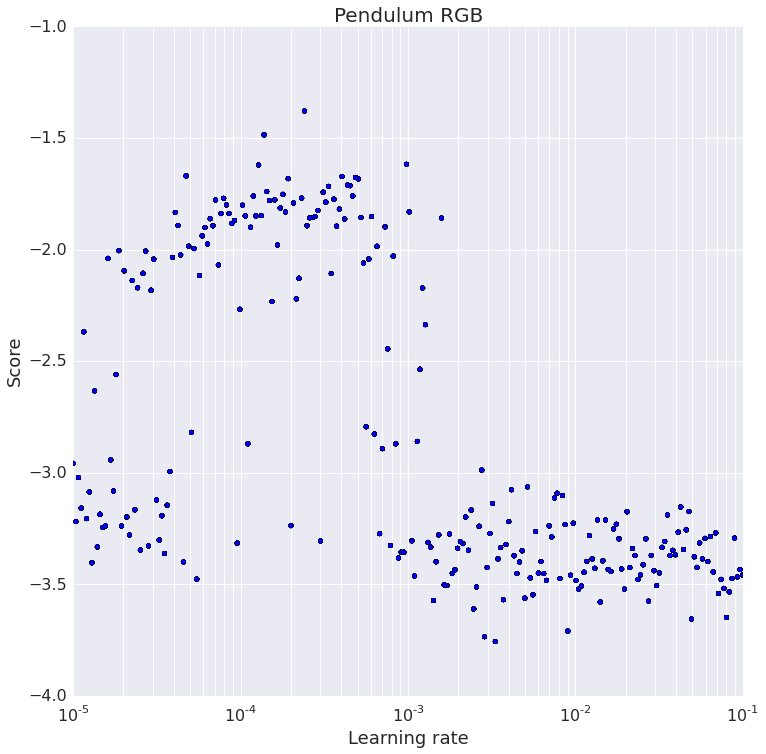}\\
    \includegraphics[width=0.24\columnwidth]{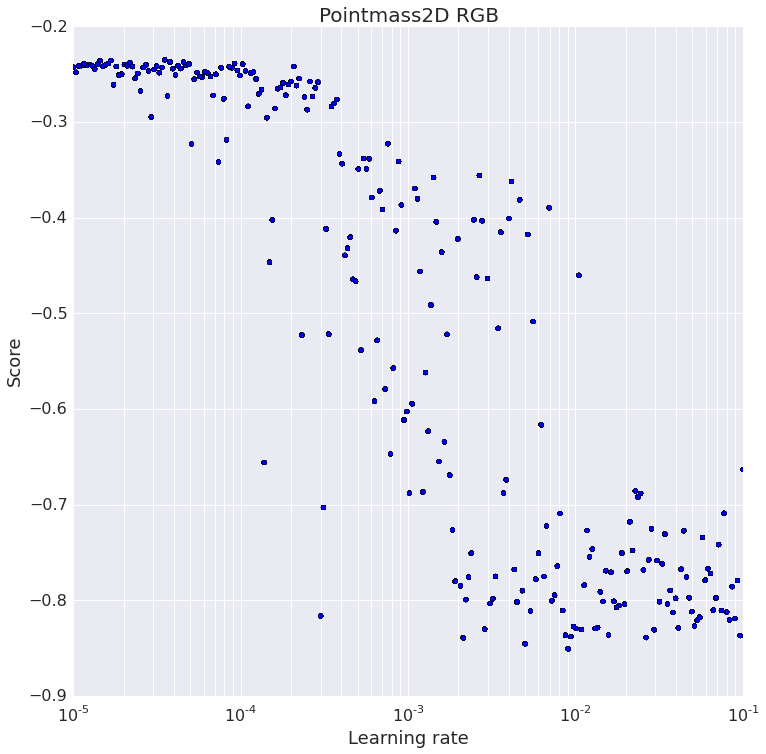}
    \includegraphics[width=0.24\columnwidth]{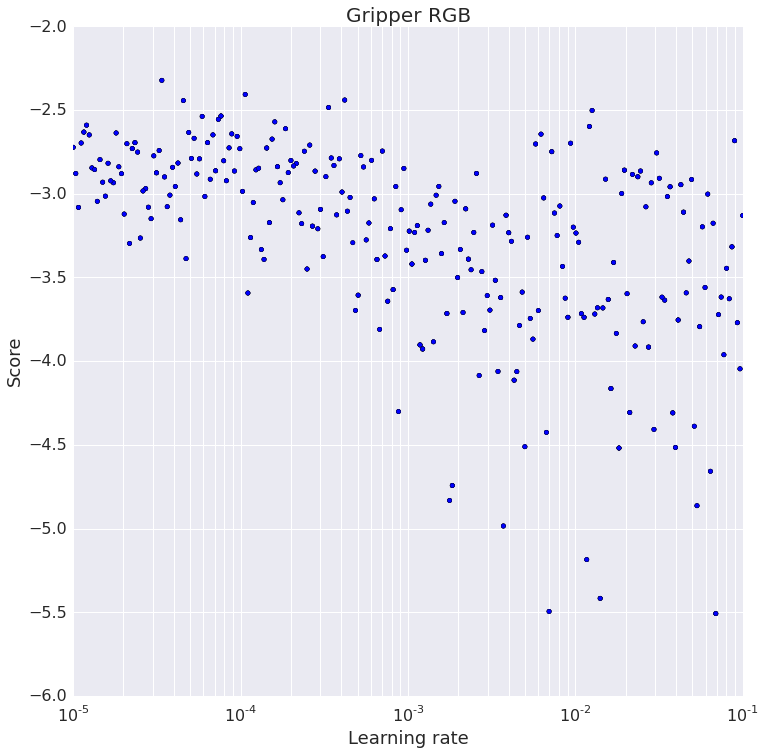}
    \caption{\label{fig-mujoco-lr} Performance for the Mujoco continuous action domains.
        Scatter plot of the best score obtained against learning rates
        sampled from $LogUniform(10^{-5},10^{-1})$.  For nearly all of the tasks
    there is a wide range of learning rates that lead to good performance on the task.}
\end{figure}

\begin{figure}[t]
    \includegraphics[width=0.24\columnwidth]{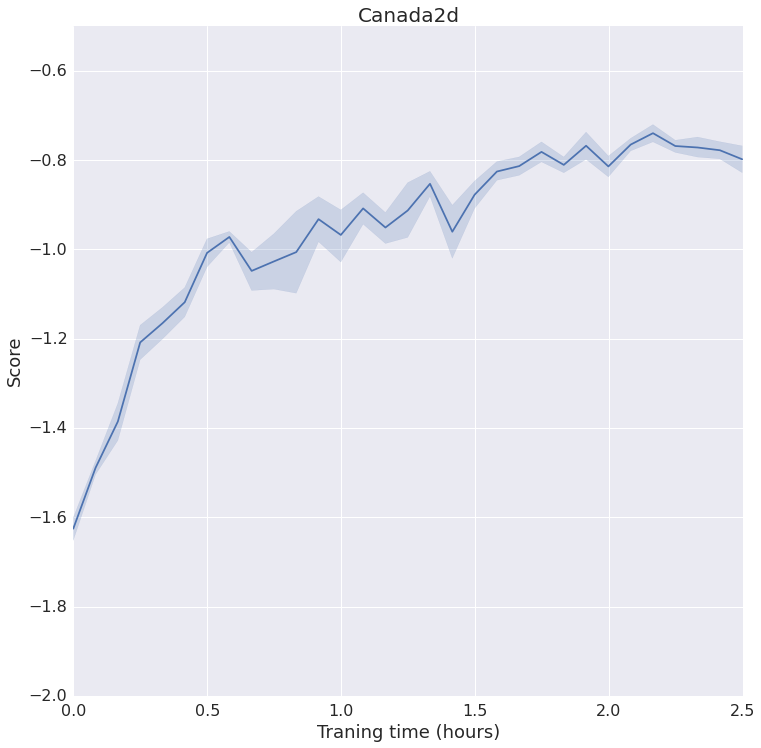}
    \includegraphics[width=0.24\columnwidth]{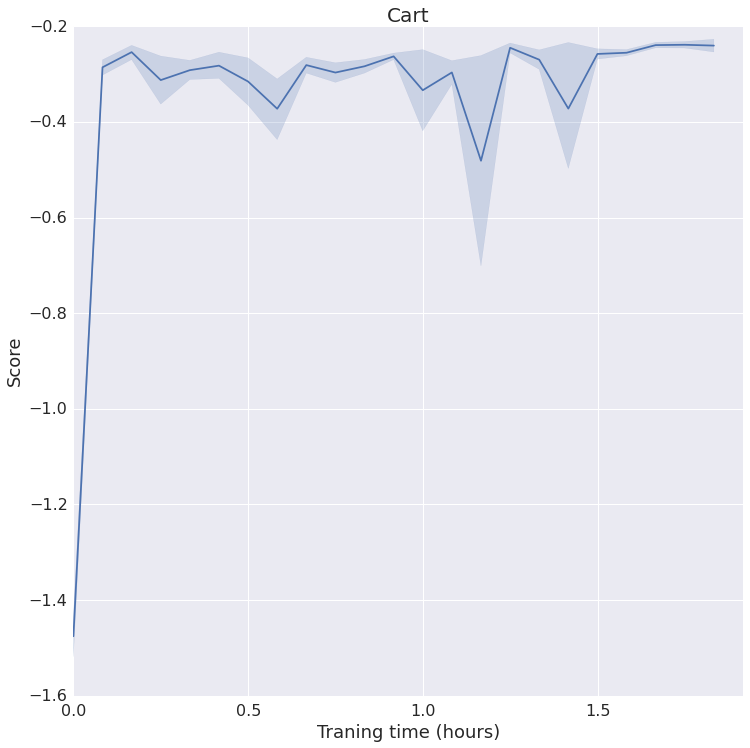}
    \includegraphics[width=0.24\columnwidth]{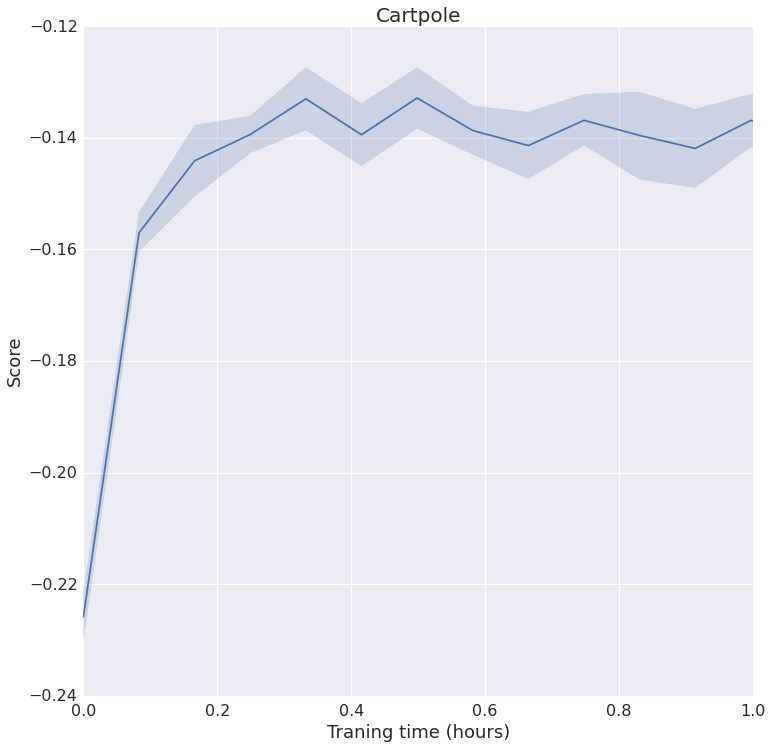}
    \includegraphics[width=0.24\columnwidth]{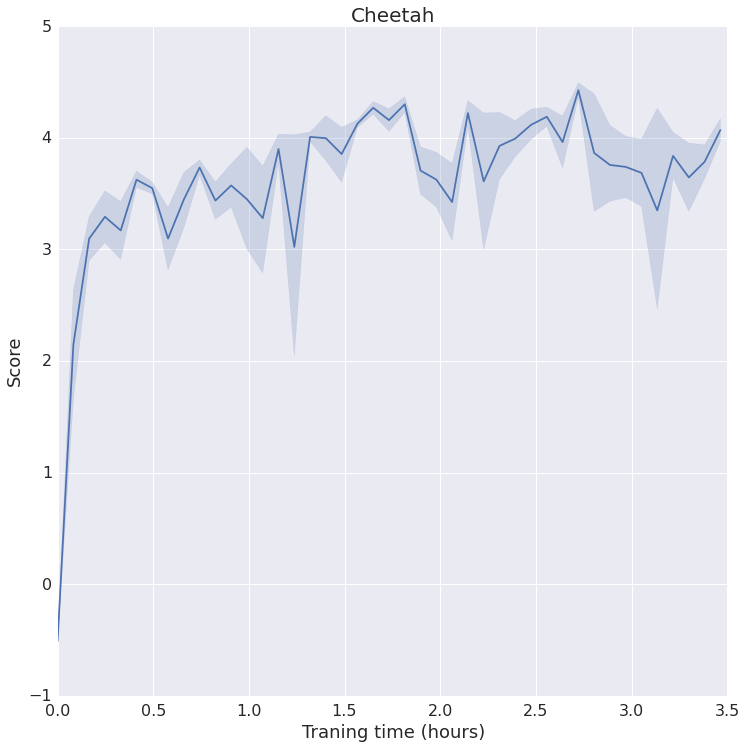}\\
    \includegraphics[width=0.24\columnwidth]{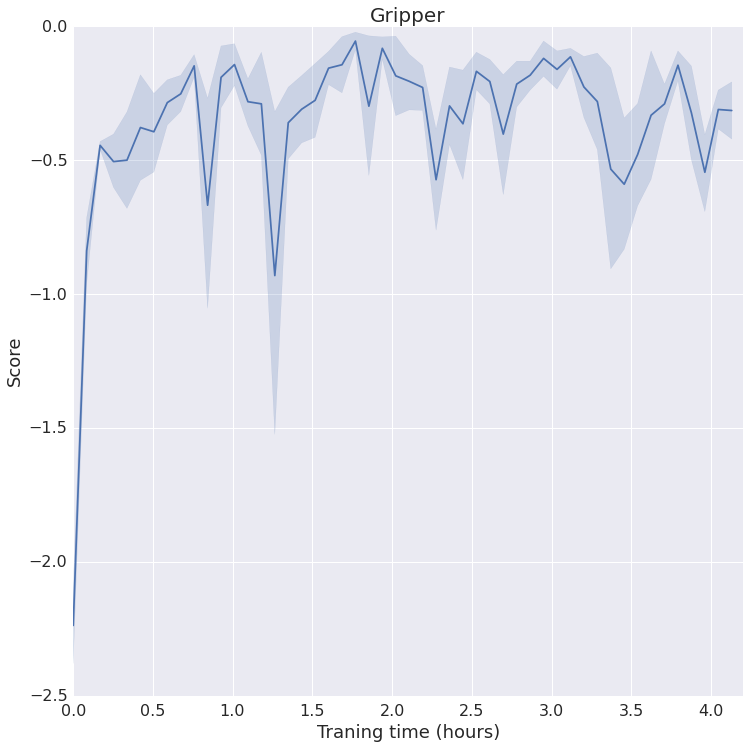}
    \includegraphics[width=0.24\columnwidth]{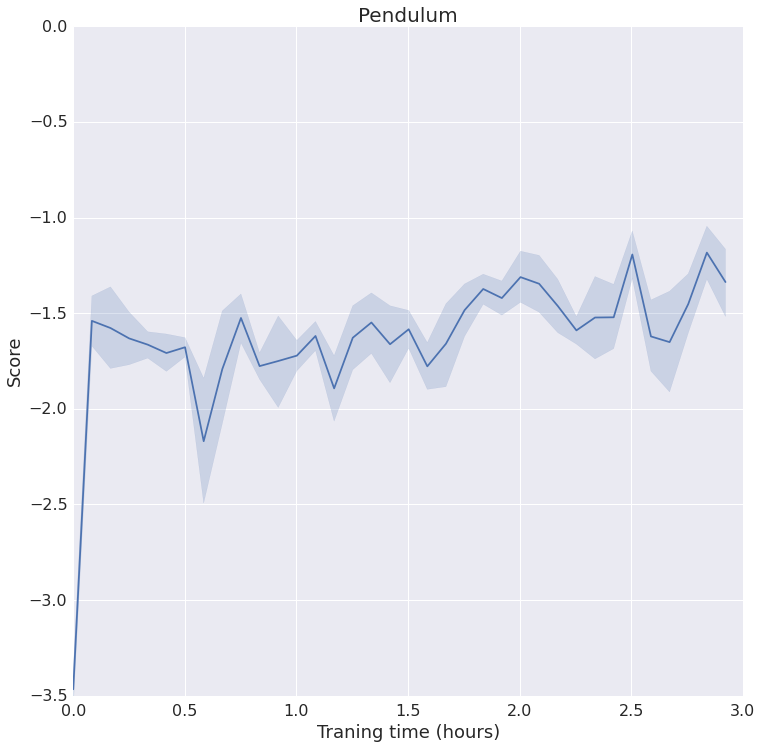}
    \includegraphics[width=0.24\columnwidth]{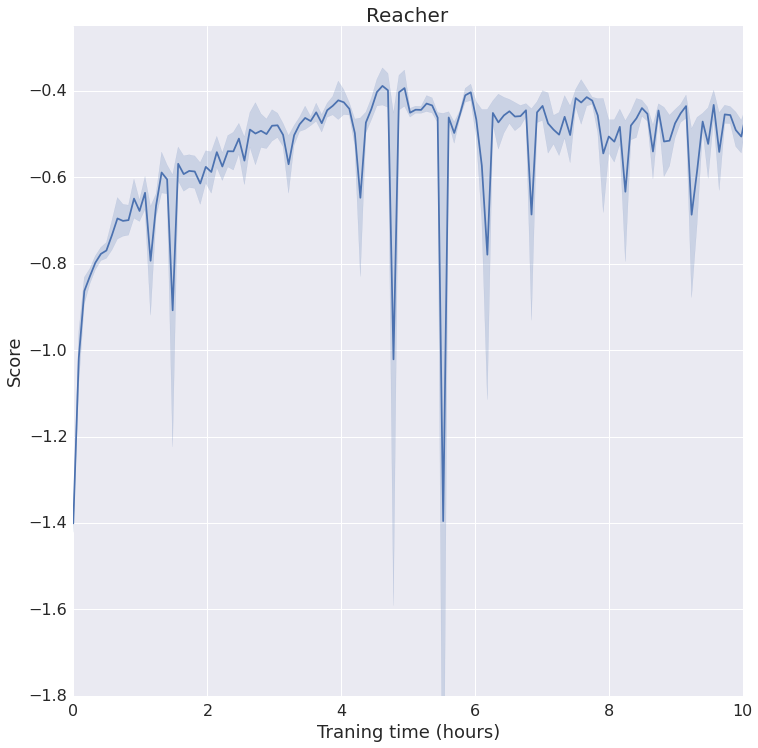}
    \includegraphics[width=0.24\columnwidth]{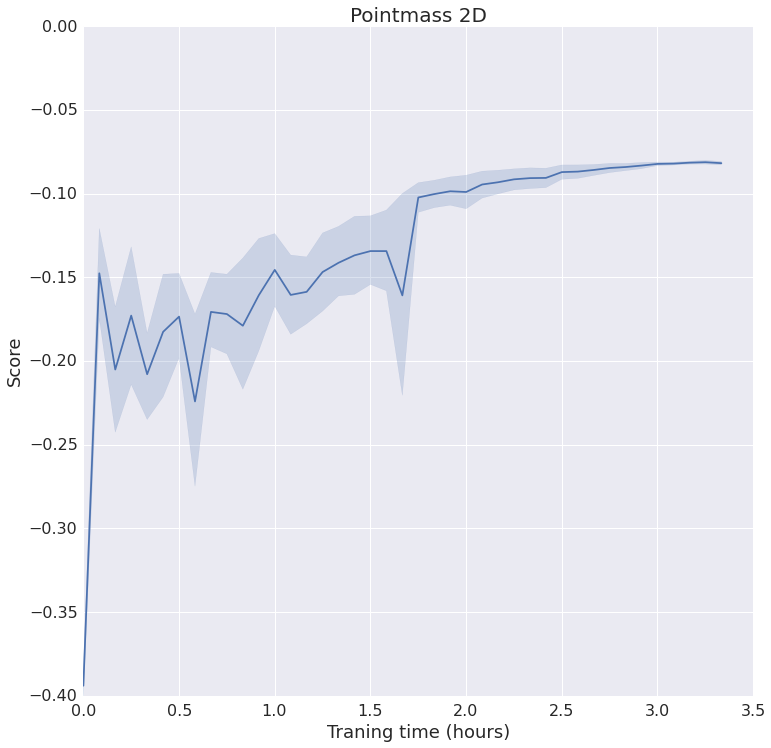}\\
    \includegraphics[width=0.24\columnwidth]{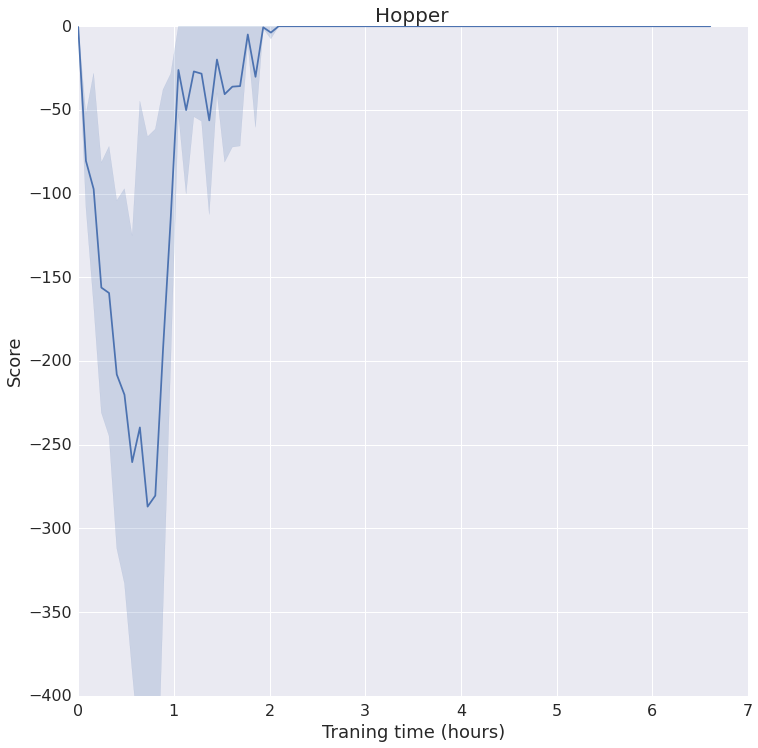}
    \includegraphics[width=0.24\columnwidth]{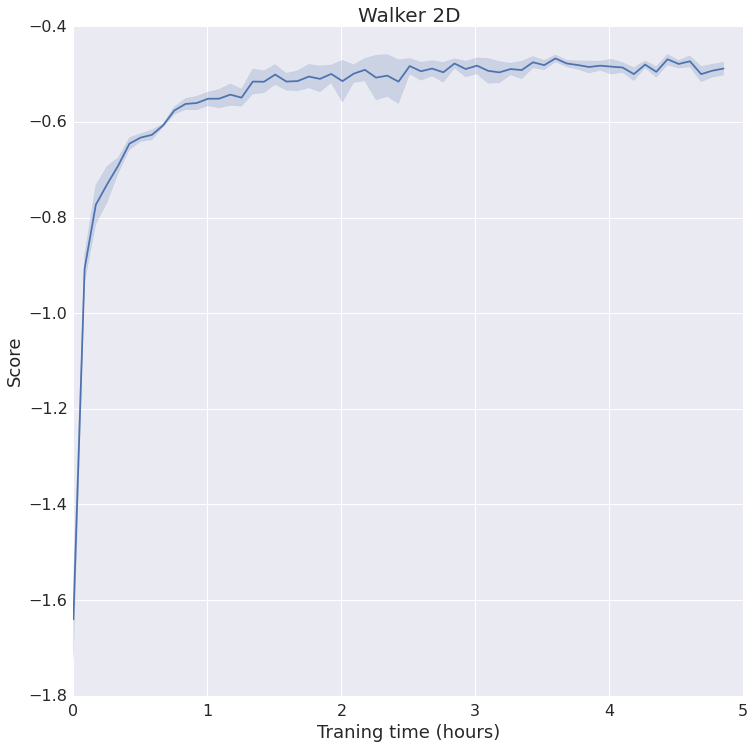}
    \includegraphics[width=0.24\columnwidth]{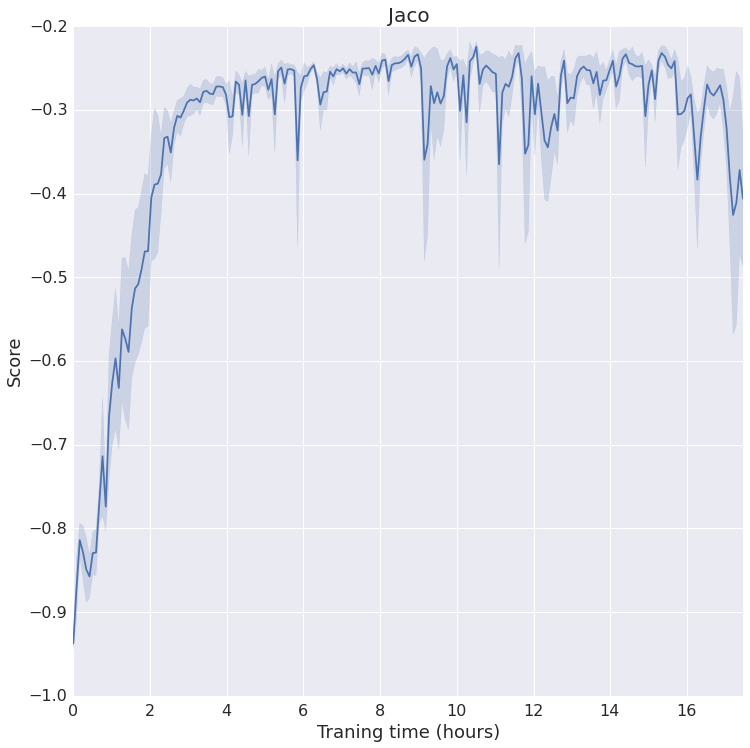}
    \includegraphics[width=0.24\columnwidth]{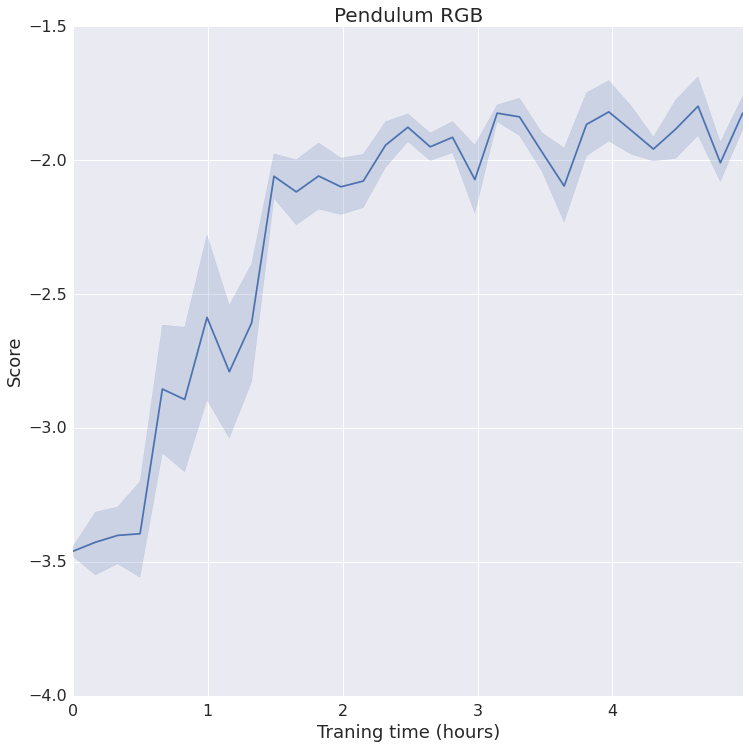}\\
    \includegraphics[width=0.24\columnwidth]{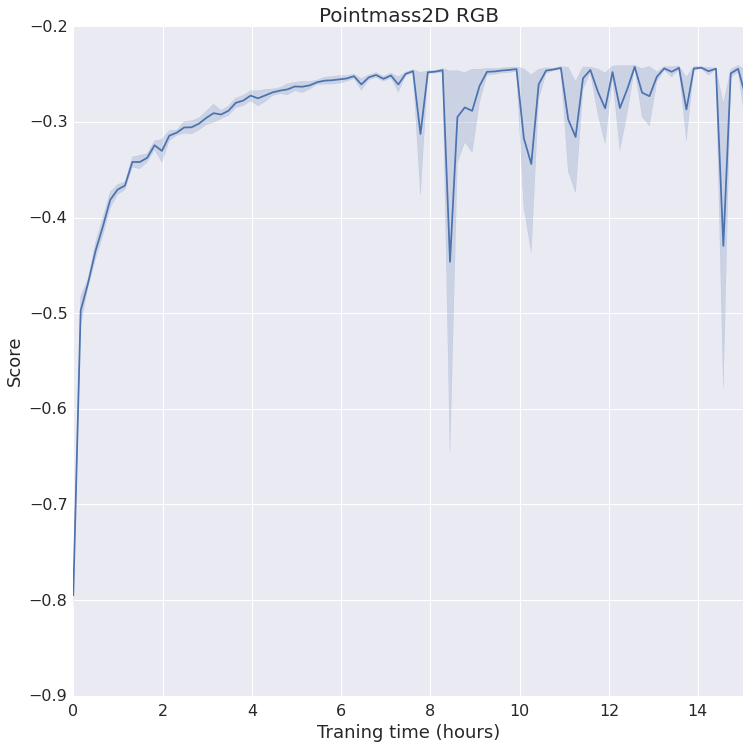}
    \includegraphics[width=0.24\columnwidth]{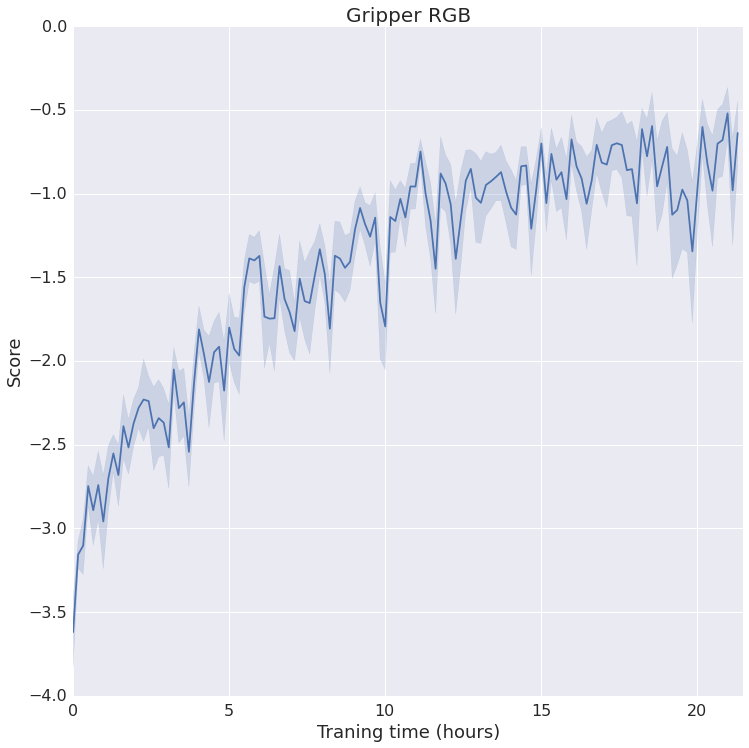}
    \caption{\label{fig-mujoco-wallclock} Score per episode vs wall-clock time
    plots for the Mujoco domains. Each plot shows error bars for the top 5 experiments.}
\end{figure}

As in the discrete action case, we included an entropy cost which encouraged exploration.
In the continuous case the we used a cost on the differential entropy of the normal distribution defined by
the output of the actor network, $-\frac{1}{2}(\log(2\pi\sigma^2)+1)$, we used a constant multiplier of $10^{-4}$ for this cost across all of the tasks
examined.
The asynchronous advantage actor-critic algorithm finds solutions for all the domains.  Figure \ref{fig-mujoco-wallclock}
shows learning curves against wall-clock time, and demonstrates that most of the domains from
states can be solved within a few hours. All of the experiments, including those done
from pixel based observations, were run on CPU.  Even in the case of solving the domains
directly from pixel inputs we found that it was possible to reliably discover solutions
within 24 hours.
Figure \ref{fig-mujoco-lr} shows scatter plots of the top scores against the sampled learning rates.
In most of the domains there is large range of learning rates that consistently achieve good performance
on the task.

\begin{algorithm}[h]
\caption{Asynchronous n-step Q-learning - pseudocode for each actor-learner thread.}
\begin{algorithmic}
\small
\State \emph{// Assume global shared parameter vector $\theta$.}
\State \emph{// Assume global shared target parameter vector $\theta^-$.}
\State \emph{// Assume global shared counter $T=0$.}
\State Initialize thread step counter $t\gets 1$
\State Initialize target network parameters $\theta^- \gets \theta$
\State Initialize thread-specific parameters $\theta' = \theta$
\State Initialize network gradients $d\theta \gets 0$
\Repeat
\State Clear gradients $d\theta \gets 0$
\State Synchronize thread-specific parameters $\theta'=\theta$
\State $t_{start} = t$
\State Get state $s_t$
\Repeat
\State Take action $a_t$ according to the $\epsilon$-greedy policy based on $Q(s_t,a;\theta')$
\State Receive reward $r_t$ and new state $s_{t+1}$
\State $t \gets t + 1$
\State $T \gets T + 1$
\Until terminal $s_t$ \textbf{or} $t - t_{start} == t_{max}$
\State $R =
    \left\{
    \begin{array}{l l}
      0  \quad & \text{for terminal } s_t\\
      \max_a Q(s_t,a;\theta^-) \quad & \text{for non-terminal } s_t
    \end{array} \right.$
\For {$i \in \{t-1,\ldots,t_{start} \}$}
\State $R \gets r_i + \gamma R$
\State Accumulate gradients wrt $\theta'$: $d\theta \gets d\theta + \frac{\partial\left(R - Q(s_i,a_i;\theta')\right)^2}{\partial \theta'}$
\EndFor
\State Perform asynchronous update of $\theta$ using $d\theta$.
\If {$T\mod I_{target} == 0$}
\State $\theta^- \gets \theta$
\EndIf
\Until $T > T_{max}$
\end{algorithmic}
\label{alg:msq}
\end{algorithm}

\begin{algorithm}[h]
\caption{Asynchronous advantage actor-critic - pseudocode for each actor-learner thread.}
\begin{algorithmic}
\small
\State \emph{// Assume global shared parameter vectors $\theta$ and $\theta_v$ and global shared counter $T=0$}
\State \emph{// Assume thread-specific parameter vectors $\theta'$ and $\theta'_v$}
\State Initialize thread step counter $t\gets 1$
\Repeat
\State Reset gradients: $d\theta \gets 0$ and $d\theta_v \gets 0$.
\State Synchronize thread-specific parameters  $\theta'=\theta$ and $\theta'_v=\theta_v$ %
\State $t_{start} = t$
\State Get state $s_t$
\Repeat
\State Perform $a_t$ according to policy $\pi (a_t|s_t;\theta')$
\State Receive reward $r_t$ and new state $s_{t+1}$
\State $t \gets t + 1$
\State $T \gets T + 1$
\Until terminal $s_t$ \textbf{or} $t-t_{start}==t_{max}$
\State $R =
    \left\{
    \begin{array}{l l}
      0  \quad & \text{for terminal } s_t\\
        V(s_t,\theta'_v) \quad & \text{for non-terminal } s_t \text{// Bootstrap from last state}
    \end{array}\right.$
\For {$i \in \{t-1,\ldots,t_{start} \}$}
\State $R \gets r_i + \gamma R$
\State Accumulate gradients wrt $\theta'$: $d\theta \gets d\theta + \nabla_{\theta'} \log\pi(a_i|s_i;\theta') (R - V(s_i;\theta'_v))$
\State Accumulate gradients wrt $\theta'_v$: $d\theta_v \gets d\theta_v + {\partial\left(R - V(s_i;\theta'_v)\right)^2}/{\partial \theta'_v}$
\EndFor
\State Perform asynchronous update of $\theta$ using $d\theta$ and of $\theta_v$ using $d\theta_v$.
\Until $T > T_{max}$
\end{algorithmic}
\label{alg:reinforce}
\end{algorithm}

\clearpage

\clearpage

\begin{figure}
\centerline{\includegraphics[width=\columnwidth]{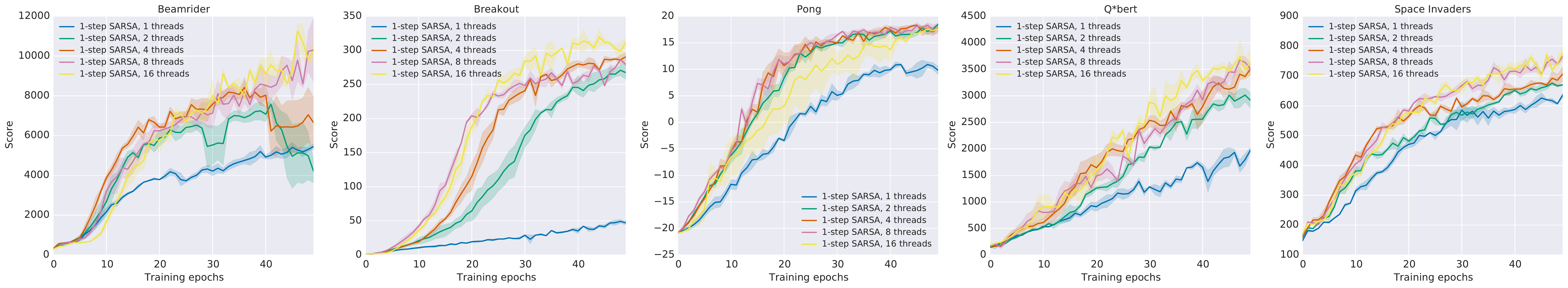}}
\caption{\label{fig-scalability-data-sarsa} Data efficiency comparison of different numbers of actor-learners one-step Sarsa on five Atari games. The x-axis shows the total number of training epochs where an epoch corresponds to four million frames (across all threads).  The y-axis shows the average score. Each curve shows the average of the three best performing agents from a search over 50 random learning rates. Sarsa shows increased data efficiency with increased numbers of parallel workers.}
\end{figure}

\begin{figure}
\centerline{\includegraphics[width=\columnwidth]{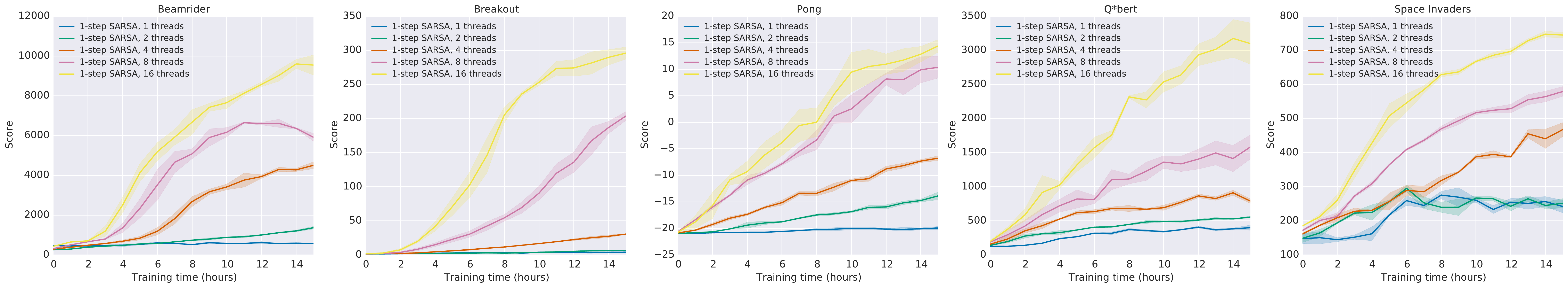}}
\caption{\label{fig-scalability-time-sarsa} Training speed comparison of different numbers of actor-learners for all one-step Sarsa on five Atari games. The x-axis shows training time in hours while the y-axis shows the average score.  Each curve shows the average of the three best performing agents from a search over 50 random learning rates.  Sarsa shows significant speedups from using greater numbers of parallel actor-learners.}
\end{figure}

\begin{figure}[t]
\centerline{\includegraphics[width=\columnwidth]{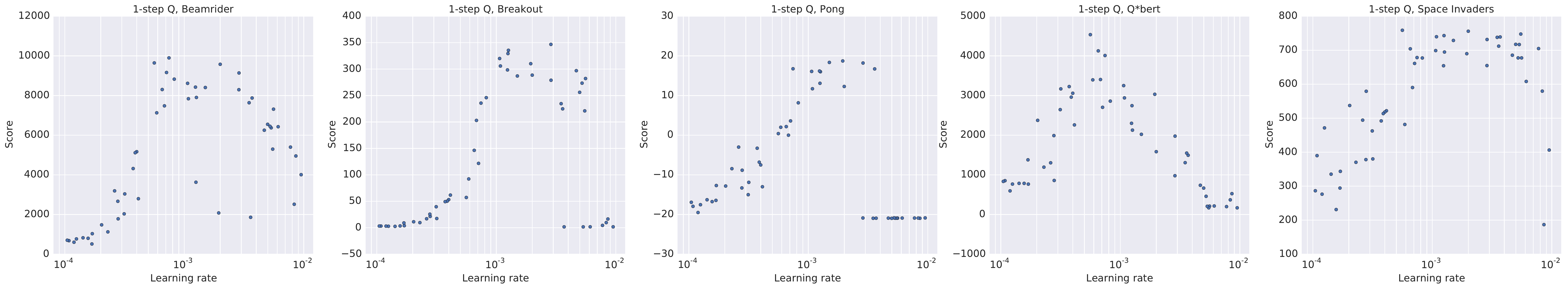}}
\centerline{\includegraphics[width=\columnwidth]{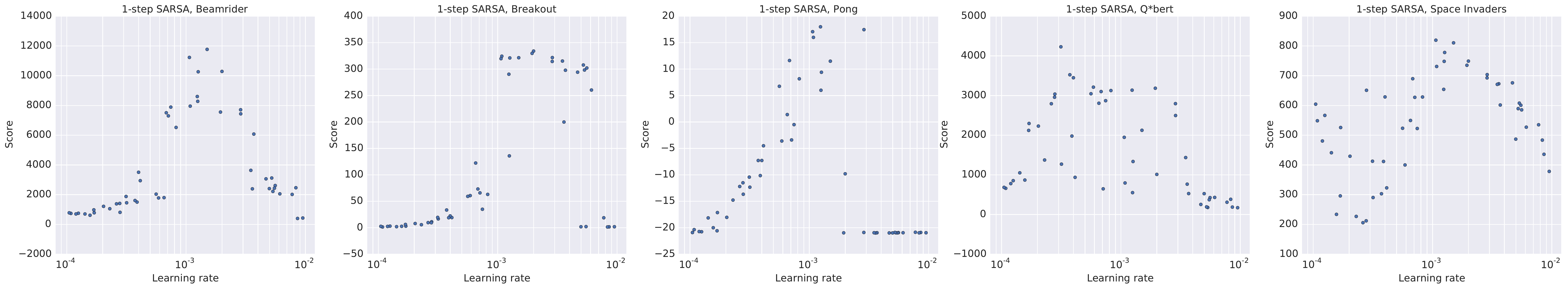}}
\centerline{\includegraphics[width=\columnwidth]{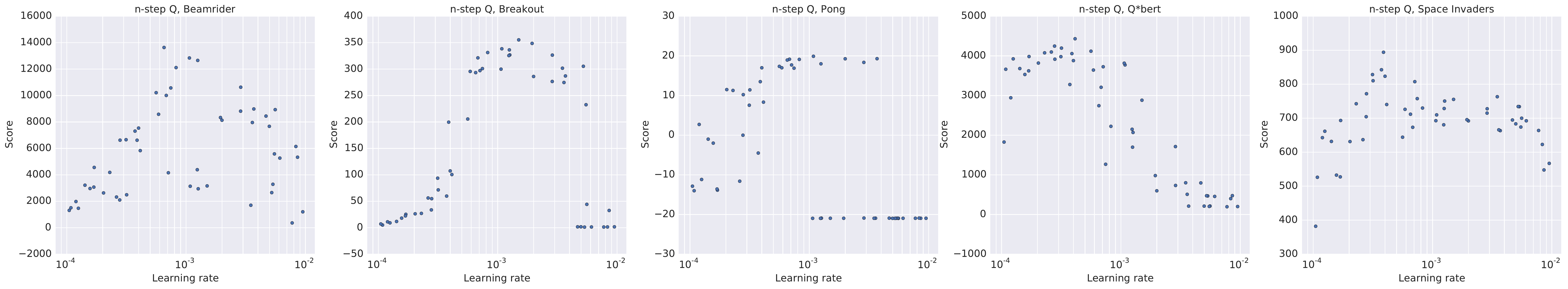}}
\caption{\label{fig-stability-lr} Scatter plots of scores obtained by one-step Q, one-step Sarsa, and $n$-step Q on five games (Beamrider, Breakout, Pong, Q*bert, Space Invaders) for $50$ different learning rates and random initializations. All algorithms exhibit some level of robustness to the choice of learning rate.}
\end{figure}

\begin{table*}[h]
    \centering
    \begin{scriptsize}
    \begin{tabular}{lrrrrrrrr}
        \textbf{Game} & \textbf{DQN} & \textbf{Gorila} & \textbf{Double} & \textbf{Dueling} & \textbf{Prioritized} & \textbf{A3C FF, 1 day} & \textbf{A3C FF } & \textbf{A3C LSTM } \\
Alien & 570.2 & 813.5 & 1033.4 & \textbf{1486.5} & 900.5 & 182.1 & 518.4 & 945.3\\
Amidar & 133.4 & 189.2 & 169.1 & 172.7 & 218.4 & \textbf{283.9} & 263.9 & 173.0\\
Assault & 3332.3 & 1195.8 & 6060.8 & 3994.8 & 7748.5 & 3746.1 & 5474.9 & \textbf{14497.9}\\
Asterix & 124.5 & 3324.7 & 16837.0 & 15840.0 & \textbf{31907.5} & 6723.0 & 22140.5 & 17244.5\\
Asteroids & 697.1 & 933.6 & 1193.2 & 2035.4 & 1654.0 & 3009.4 & 4474.5 & \textbf{5093.1}\\
Atlantis & 76108.0 & 629166.5 & 319688.0 & 445360.0 & 593642.0 & 772392.0 & \textbf{911091.0} & 875822.0\\
Bank Heist & 176.3 & 399.4 & 886.0 & \textbf{1129.3} & 816.8 & 946.0 & 970.1 & 932.8\\
Battle Zone & 17560.0 & 19938.0 & 24740.0 & \textbf{31320.0} & 29100.0 & 11340.0 & 12950.0 & 20760.0\\
Beam Rider & 8672.4 & 3822.1 & 17417.2 & 14591.3 & \textbf{26172.7} & 13235.9 & 22707.9 & 24622.2\\
Berzerk &  &  & 1011.1 & 910.6 & 1165.6 & \textbf{1433.4} & 817.9 & 862.2\\
Bowling & 41.2 & 54.0 & \textbf{69.6} & 65.7 & 65.8 & 36.2 & 35.1 & 41.8\\
Boxing & 25.8 & 74.2 & 73.5 & \textbf{77.3} & 68.6 & 33.7 & 59.8 & 37.3\\
Breakout & 303.9 & 313.0 & 368.9 & 411.6 & 371.6 & 551.6 & 681.9 & \textbf{766.8}\\
Centipede & 3773.1 & \textbf{6296.9} & 3853.5 & 4881.0 & 3421.9 & 3306.5 & 3755.8 & 1997.0\\
Chopper Comman & 3046.0 & 3191.8 & 3495.0 & 3784.0 & 6604.0 & 4669.0 & 7021.0 & \textbf{10150.0}\\
Crazy Climber & 50992.0 & 65451.0 & 113782.0 & 124566.0 & 131086.0 & 101624.0 & 112646.0 & \textbf{138518.0}\\
Defender &  &  & 27510.0 & 33996.0 & 21093.5 & 36242.5 & 56533.0 & \textbf{233021.5}\\
Demon Attack & 12835.2 & 14880.1 & 69803.4 & 56322.8 & 73185.8 & 84997.5 & 113308.4 & \textbf{115201.9}\\
Double Dunk & -21.6 & -11.3 & -0.3 & -0.8 & \textbf{2.7} & 0.1 & -0.1 & 0.1\\
Enduro & 475.6 & 71.0 & 1216.6 & \textbf{2077.4} & 1884.4 & -82.2 & -82.5 & -82.5\\
Fishing Derby & -2.3 & 4.6 & 3.2 & -4.1 & 9.2 & 13.6 & 18.8 & \textbf{22.6}\\
Freeway & 25.8 & 10.2 & \textbf{28.8} & 0.2 & 27.9 & 0.1 & 0.1 & 0.1\\
Frostbite & 157.4 & 426.6 & 1448.1 & 2332.4 & \textbf{2930.2} & 180.1 & 190.5 & 197.6\\
Gopher & 2731.8 & 4373.0 & 15253.0 & 20051.4 & \textbf{57783.8} & 8442.8 & 10022.8 & 17106.8\\
Gravitar & 216.5 & \textbf{538.4} & 200.5 & 297.0 & 218.0 & 269.5 & 303.5 & 320.0\\
H.E.R.O. & 12952.5 & 8963.4 & 14892.5 & 15207.9 & 20506.4 & 28765.8 & \textbf{32464.1} & 28889.5\\
Ice Hockey & -3.8 & -1.7 & -2.5 & -1.3 & \textbf{-1.0} & -4.7 & -2.8 & -1.7\\
James Bond & 348.5 & 444.0 & 573.0 & 835.5 & \textbf{3511.5} & 351.5 & 541.0 & 613.0\\
Kangaroo & 2696.0 & 1431.0 & \textbf{11204.0} & 10334.0 & 10241.0 & 106.0 & 94.0 & 125.0\\
Krull & 3864.0 & 6363.1 & 6796.1 & 8051.6 & 7406.5 & \textbf{8066.6} & 5560.0 & 5911.4\\
Kung-Fu Master & 11875.0 & 20620.0 & 30207.0 & 24288.0 & 31244.0 & 3046.0 & 28819.0 & \textbf{40835.0}\\
Montezuma's Revenge & 50.0 & \textbf{84.0} & 42.0 & 22.0 & 13.0 & 53.0 & 67.0 & 41.0\\
Ms. Pacman & 763.5 & 1263.0 & 1241.3 & \textbf{2250.6} & 1824.6 & 594.4 & 653.7 & 850.7\\
Name This Game & 5439.9 & 9238.5 & 8960.3 & 11185.1 & 11836.1 & 5614.0 & 10476.1 & \textbf{12093.7}\\
Phoenix &  &  & 12366.5 & 20410.5 & 27430.1 & 28181.8 & 52894.1 & \textbf{74786.7}\\
Pit Fall &  &  & -186.7 & -46.9 & \textbf{-14.8} & -123.0 & -78.5 & -135.7\\
Pong & 16.2 & 16.7 & \textbf{19.1} & 18.8 & 18.9 & 11.4 & 5.6 & 10.7\\
Private Eye & 298.2 & \textbf{2598.6} & -575.5 & 292.6 & 179.0 & 194.4 & 206.9 & 421.1\\
Q*Bert & 4589.8 & 7089.8 & 11020.8 & 14175.8 & 11277.0 & 13752.3 & 15148.8 & \textbf{21307.5}\\
River Raid & 4065.3 & 5310.3 & 10838.4 & 16569.4 & \textbf{18184.4} & 10001.2 & 12201.8 & 6591.9\\
Road Runner & 9264.0 & 43079.8 & 43156.0 & 58549.0 & 56990.0 & 31769.0 & 34216.0 & \textbf{73949.0}\\
Robotank & 58.5 & 61.8 & 59.1 & \textbf{62.0} & 55.4 & 2.3 & 32.8 & 2.6\\
Seaquest & 2793.9 & 10145.9 & 14498.0 & 37361.6 & \textbf{39096.7} & 2300.2 & 2355.4 & 1326.1\\
Skiing &  &  & -11490.4 & -11928.0 & \textbf{-10852.8} & -13700.0 & -10911.1 & -14863.8\\
Solaris &  &  & 810.0 & 1768.4 & \textbf{2238.2} & 1884.8 & 1956.0 & 1936.4\\
Space Invaders & 1449.7 & 1183.3 & 2628.7 & 5993.1 & 9063.0 & 2214.7 & 15730.5 & \textbf{23846.0}\\
Star Gunner & 34081.0 & 14919.2 & 58365.0 & 90804.0 & 51959.0 & 64393.0 & 138218.0 & \textbf{164766.0}\\
Surround &  &  & 1.9 & \textbf{4.0} & -0.9 & -9.6 & -9.7 & -8.3\\
Tennis & -2.3 & -0.7 & -7.8 & \textbf{4.4} & -2.0 & -10.2 & -6.3 & -6.4\\
Time Pilot & 5640.0 & 8267.8 & 6608.0 & 6601.0 & 7448.0 & 5825.0 & 12679.0 & \textbf{27202.0}\\
Tutankham & 32.4 & 118.5 & 92.2 & 48.0 & 33.6 & 26.1 & \textbf{156.3} & 144.2\\
Up and Down & 3311.3 & 8747.7 & 19086.9 & 24759.2 & 29443.7 & 54525.4 & 74705.7 & \textbf{105728.7}\\
Venture & 54.0 & \textbf{523.4} & 21.0 & 200.0 & 244.0 & 19.0 & 23.0 & 25.0\\
Video Pinball & 20228.1 & 112093.4 & 367823.7 & 110976.2 & 374886.9 & 185852.6 & 331628.1 & \textbf{470310.5}\\
Wizard of Wor & 246.0 & 10431.0 & 6201.0 & 7054.0 & 7451.0 & 5278.0 & 17244.0 & \textbf{18082.0}\\
Yars Revenge &  &  & 6270.6 & \textbf{25976.5} & 5965.1 & 7270.8 & 7157.5 & 5615.5\\
Zaxxon & 831.0 & 6159.4 & 8593.0 & 10164.0 & 9501.0 & 2659.0 & \textbf{24622.0} & 23519.0\\

    \end{tabular}
    \end{scriptsize}
    \caption{\label{fig-atari-full} Raw scores for the human start condition (30 minutes emulator time). DQN scores taken from~\cite{nair2015gorila}. Double DQN scores taken from~\cite{hado2015doubledqn}, Dueling scores from~\cite{wang2015dueling} and Prioritized scores taken from~\cite{schaul2015prioritized}}
\end{table*}

\end{document}